\pdfoutput=1

\documentclass[11pt]{article}

\usepackage{acl}

\usepackage{times}
\usepackage{latexsym}

\usepackage[T1]{fontenc}

\usepackage[utf8]{inputenc}

\usepackage{microtype}

\usepackage{inconsolata}

\PassOptionsToPackage{table,xcdraw}{xcolor} 
\usepackage{graphicx}

\usepackage{booktabs}
\usepackage{amssymb}
\usepackage{multirow}
\usepackage{amsmath}
\usepackage{colortbl}
\usepackage{tabularx}
\usepackage{bbm}
\usepackage{subcaption}
\usepackage{enumitem}
\usepackage{xcolor}
\usepackage{array}
\usepackage{float}
\usepackage{tikz}
\definecolor{seagreen}{RGB}{46,139,87}     
\definecolor{firebrick}{RGB}{178,34,34}    
\newcolumntype{Y}{>{\centering\arraybackslash}X}

\newcommand{\upDelta}[1]{{\scriptsize\textcolor{seagreen}{$\uparrow$#1}}}
\newcommand{\downDelta}[1]{{\scriptsize\textcolor{firebrick}{$\downarrow$#1}}}

\DeclareRobustCommand*\circled[1]{\tikz[baseline=(char.base)]{
            \node[shape=circle,draw,inner sep=1pt] (char) {#1};}}

\title{Aligning What LLMs Do and Say: Towards Self-Consistent Explanations}

\author{
  \textbf{Sahar Admoni\textsuperscript{1}},
  \textbf{Ofra Amir\textsuperscript{1}},
  \textbf{Assaf Hallak\textsuperscript{2}},
  \textbf{Yftah Ziser\textsuperscript{2,3}}
  \\
  \textsuperscript{1}Technion -- Israel Institute of Technology \quad
  \textsuperscript{2}Nvidia Research \quad
  \textsuperscript{3}University of Groningen
  \\
  \texttt{saharad@campus.technion.ac.il}, \texttt{oamir@technion.ac.il}, \texttt{\{ahallak,yziser\}@nvidia.com}
}

\begin{document}
\maketitle
\begin{abstract}
Large language models (LLMs) seem to offer an easy path to interpretability: just ask them to explain their answers. Yet the features driving an answer often differ from those emphasized in its explanation, meaning post-hoc rationales can misrepresent what actually shaped the model's output. We quantify this gap by comparing the feature-importance distributions of answers and their explanations. Prior analyses reveal such discrepancies, but large-scale study has been limited by the high computational cost of attribution methods. To address this, we introduce the Post-hoc Self-Consistency Bank (PSCB), a large-scale benchmark linking model decisions with diverse explanations and attribution vectors across datasets, methods, and model families. Using PSCB, we find that Spearman rank correlation provides a more reliable signal of alignment than cosine similarity. Building on this insight, we apply Direct Preference Optimization (DPO) to attribution-based preference data, improving alignment without degrading task accuracy, and show that standard supervised fine-tuning on the same data fails to achieve comparable gains. These improvements generalize robustly across domains, paving the way toward scalable and faithful alignment between LLM decisions and their natural language explanations.\footnote{Code and data are available at \url{https://github.com/saharad1/ConstLLM}.}

\end{abstract}

\begin{figure}[t]
    \centering
    \includegraphics[width=\columnwidth]{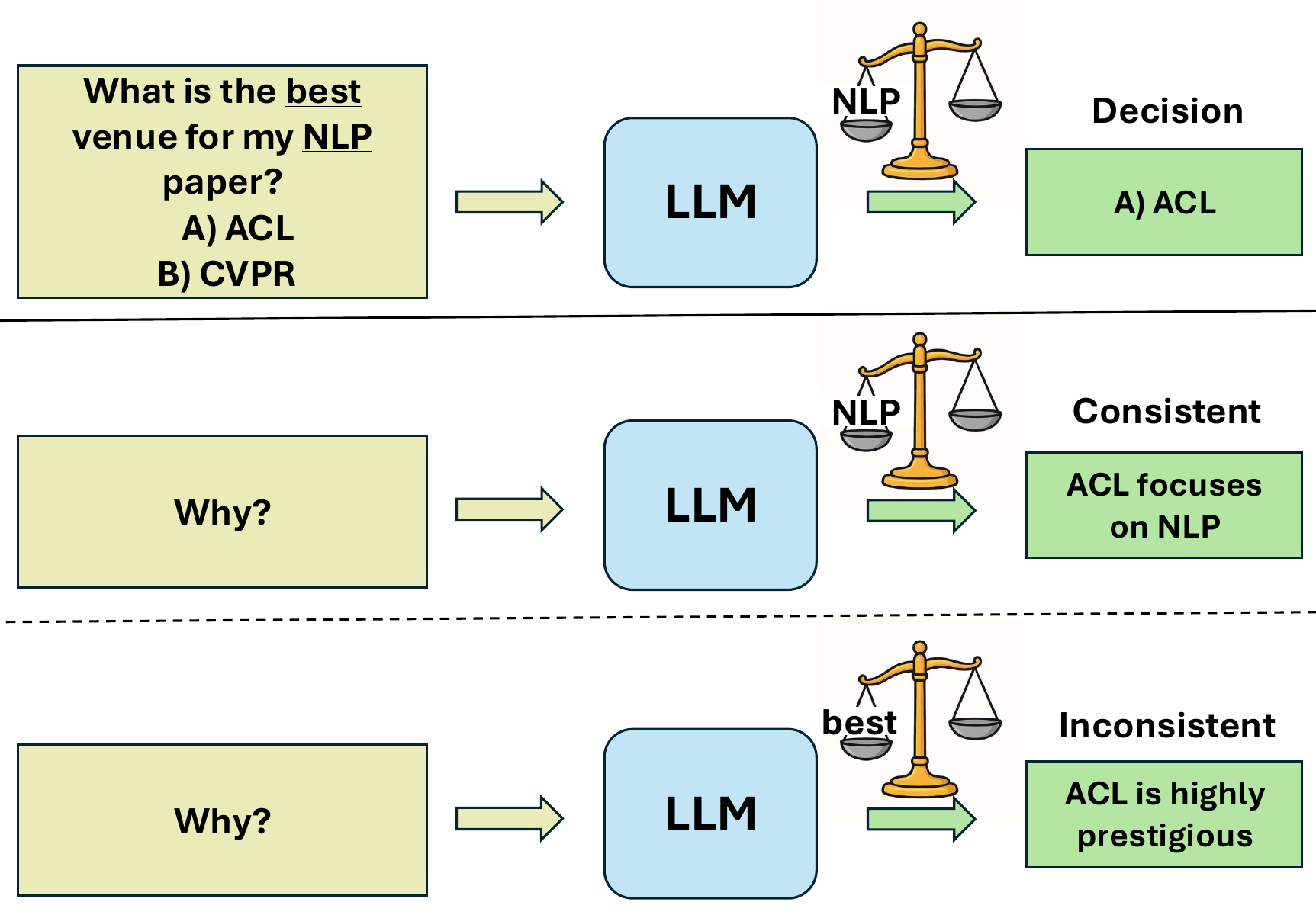}
    \caption{Illustration of explanation consistency using feature importance. \textbf{Top:} The LLM generates an answer where the word \emph{NLP} in the prompt has high feature importance. \textbf{Middle:} A consistent explanation where \emph{NLP} in the prompt also has high feature importance. \textbf{Bottom:} An inconsistent explanation, where the word \emph{best} in the prompt has high feature importance instead, misaligned with the model’s actual decision.}
    \label{fig:main_intro}
\end{figure}

\section{Introduction}
As LLMs are increasingly embedded in user-facing and decision-support systems, their outputs and accompanying explanations are often trusted by end users even when independent verification is impractical~\cite{sun2024trustllm,liu2023trustworthy,doshi2017towards}. To foster such trust, LLMs are frequently prompted to produce natural language explanations~\cite{madsen2024self}. Yet these post-hoc explanations often draw on different input features than those that determined the model’s answer~\cite{turpin2023language,randl2024evaluating,hase2020leakage}, revealing a gap we term \emph{self-consistency}. While faithfulness broadly concerns whether explanations reflect the model’s decision-making process~\cite{jacovi2020towards,wiegreffe2020measuring}, prior work suggests that many proposed faithfulness metrics in fact capture forms of self-consistency, agreement between the factors influencing a model’s prediction and those invoked in its explanation~\cite{parcalabescu2023measuring}. We operationalize this notion through feature-attribution alignment, measuring whether the explanation relies on the same input evidence that influenced the model’s answer. Our work focuses on quantifying and improving this property for post-hoc explanations generated after the answer (see Figure~\ref{fig:main_intro}).

Prior works apply counterfactual interventions to input features using various heuristics to estimate their influence on model decisions~\cite{wiegreffe2020measuring,turpin2023language,lanham2023measuring,atanasova2023faithfulness}. However, applying such interventions to LLMs is computationally expensive. \citet{parcalabescu2023measuring} recently proposed a more rigorous approach based on feature importance, where self-consistency is defined as the similarity between the importance assigned to the answer and to its explanation. Yet, computing feature importance itself is resource-intensive, and their evaluation was limited to only 100 test examples, constraining the conclusions that can be drawn. To the best of our knowledge, no prior work has demonstrated how to improve this alignment, as even measuring it remains highly expensive.

We address this gap with the following contributions:
(1)~\textbf{Post-hoc Self-Consistency Bank (PSCB):} a large-scale benchmark linking over 85{,}000 decisions with 428{,}000 explanations and attribution vectors (LIME and LIG) across four QA datasets and two LLMs, enabling systematic evaluation of explanation–decision alignment.
(2)~\textbf{Empirical analysis:} The first large-scale study of attribution-based self-consistency, showing it is largely orthogonal to correctness and that Spearman rank correlation provides a more discriminative signal than cosine similarity, effectively separating high- and low-quality explanations.
(3)~\textbf{Preference-based optimization:} Attribution-based preference data from PSCB is used to fine-tune LLMs with DPO, yielding substantial in-domain gains and robust cross-domain generalization without degrading accuracy.
(4)~\textbf{Multidimensionality:} Improvements transfer across domains but not across attribution paradigms, revealing that different methods capture fundamentally distinct notions of input relevance, with direct implications for how the community evaluates attribution-based explanations.

\begin{figure*}[t]
    \centering
    \includegraphics[width=\linewidth]{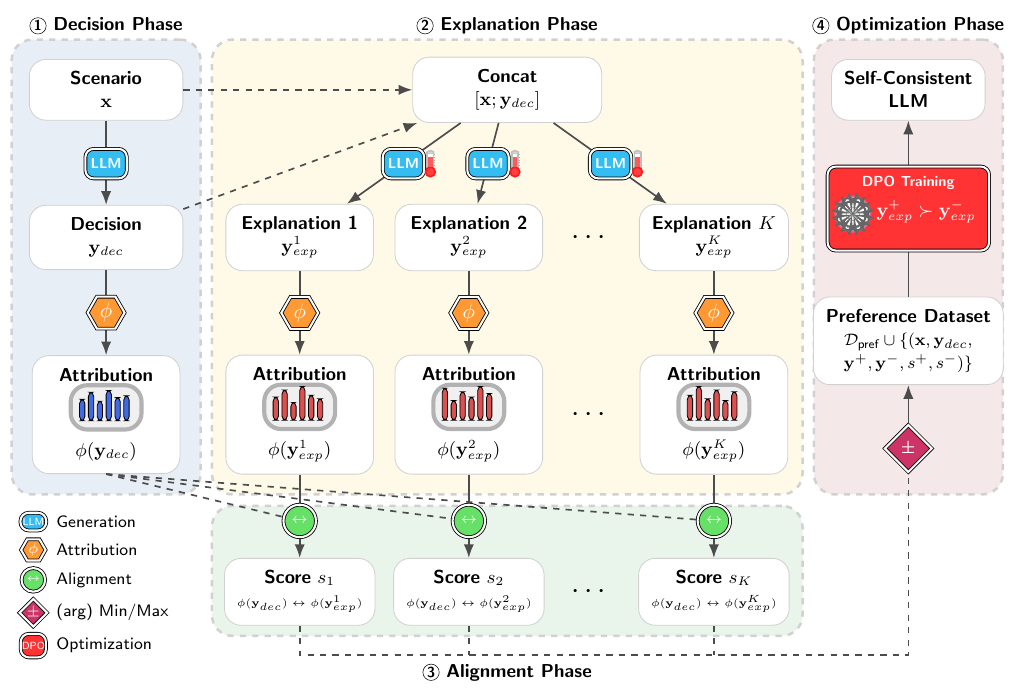}
    \caption{Overview of the PSCB pipeline. \circled{1} \textbf{Decision:} Given a multiple-choice QA instance, the LLM generates an answer together with attribution scores over the input. \circled{2} \textbf{Explanation:} The model is then prompted to produce $K$ diverse explanations (via temperature sampling) conditioned on the question and answer, with attribution scores computed for each explanation. \circled{3} \textbf{Alignment:} Self-consistency scores are obtained by comparing attribution vectors from the decision and explanations. \circled{4} \textbf{Optimization:} Finally, attribution-based preference pairs are used to fine-tune the model DPO, yielding more self-consistent explanations.}
    \label{fig:example}
\end{figure*}
\section{Background and Related Work}
Feature attribution methods estimate how input components (e.g., tokens) or internal elements (e.g., attention heads) influence model predictions~\cite{Zhao2023ExplainabilityFL}. \emph{Attention-based} methods are simple but often unreliable~\citep{jain2019attention,serrano2019attention,wiegreffe2019attention}, whereas \emph{gradient-} and \emph{perturbation-based} methods provide stronger signals, including Integrated Gradients~\cite{sundararajan2017axiomatic}, LRP~\cite{bach2015pixel,montavon2019layer}, SHAP~\cite{lundberg2017unified}, and LIME~\cite{ribeiro2016should}. SHAP is rigorous but slow; LIME trades some precision for efficiency. 

Despite advances in robustness and faithfulness~\cite{parcalabescu2023measuring,atanasova2023faithfulness}, scalable, model-agnostic attribution remains challenging. LLMs often produce fluent yet unfaithful explanations~\citep{narang2020wt5,turpin2023language,madsen2024self}, motivating \emph{self-consistency}, alignment between features driving predictions and  explanations. Perturbation-based tests evaluate robustness, while attribution-based methods compare importance vectors directly. Prior work such as CC-SHAP~\citep{parcalabescu2023measuring} handles only $\sim$100 samples; our Post-hoc Self-Consistency Bank (PSCB) scales this to tens of thousands via a LIME-based variant (CC-LIME). We also explore Layer-Integrated Gradients, a gradient-based method with cost between LIME and SHAP. Complementary efforts enhance explanation fidelity via probability-aware metrics~\citep{siegel2024probabilities}, context-faithful prompting~\citep{zhou2023context}, or cross-example semantic coherence~\citep{chen2024towards}. Other recent work targets related but distinct objectives: PEX~\citep{zhao2025pex} evaluates whether an explanation linguistically supports a label by comparing label-explanation log-odds, and CCT~\citep{siegel2024probabilities} computes population-level correlations between counterfactual interventions and explanation mentions. Both operate at the level of probabilistic or semantic consistency rather than input-level attribution alignment. Because these methods do not access or compare token-level attribution vectors, they measure a complementary dimension of explanation quality; a model can score well on probabilistic consistency while still exhibiting misaligned feature attributions between its decision and explanation. Our work focuses on \textit{attributional} self-consistency within each decision, directly comparing the input evidence used for the answer with that invoked in the explanation, and provides the first scalable framework for both measuring and improving this property.

\section{Post-hoc Self-Consistency Bank (PSCB)}

We present the \textit{Post-hoc Self-Consistency Bank} (PSCB), a collection of attribution-augmented QA datasets for evaluating decision–explanation alignment. Each dataset is defined by three components: a base QA dataset, a target LLM, and an alignment metric. We describe our construction choices and report key statistics and findings. The pipeline (Figure~\ref{fig:example}) is agnostic to the choice of the attribution method and alignment metric.

\subsection{Sequence Feature Attribution}
\label{sec:seq_attr}

We compute token-level attribution scores to analyze how input tokens influence a model’s generation. Given an input $\mathbf{x} = [x_1,\dots,x_m]$ and an output $\mathbf{y} = [y_1,\dots,y_n]$, the model generates $\mathbf{y}$ autoregressively with logits $\ell_t \in \mathbb{R}^V$ at each step $t$. We define the \textit{sequence log-probability} as
$\mathrm{SLP}(\mathbf{x}, \mathbf{y}) = \sum_{t=1}^{n} \log P\!\left(y_t \mid \mathbf{x}, \mathbf{y}_{<t}\right)$,
where $P\!\left(y_t \mid \mathbf{x}, \mathbf{y}_{<t}\right)
= \frac{\exp(\ell_t[y_t])}{\sum_{v \in V} \exp(\ell_t[v])}$,
and $\ell_t$ denotes the model logits at step $t$ conditioned on
$(\mathbf{x}, \mathbf{y}_{<t})$. This scalar serves as the attribution target, indicating how much each token in $\mathbf{x}$ contributes to generating $\mathbf{y}$.
For each output $\mathbf{y}$, we compute an attribution vector 
$\boldsymbol{\phi}^{(\mathbf{y})} = [\phi_1^{(\mathbf{y})}, \dots, \phi_m^{(\mathbf{y})}]$,
where $\phi_i^{(\mathbf{y})}$ quantifies the contribution of $x_i$. The attribution method, gradient-based, perturbation-based, or sampling-based, determines how $\boldsymbol{\phi}^{(\mathbf{y})}$ is computed; our framework is method-agnostic (Section~\ref{sec:exp_setup}). To reduce noise, we exclude a set of \emph{skip tokens} $\mathcal{S} \subseteq V$ (e.g., punctuation, formatting markers), ensuring attribution focuses on semantically meaningful inputs.

\subsection{Measuring Self-Consistency}
\label{sec:att_alig}

Building on the attribution vectors defined above, we evaluate whether an explanation reflects the reasoning behind a decision. For an input $\mathbf{x}$ with decision $\mathbf{y}_{\mathrm{dec}}$ and explanation $\mathbf{y}_{\mathrm{exp}}$, let $\boldsymbol{\phi}^{(\mathbf{y}_{\mathrm{dec}})}$ and $\boldsymbol{\phi}^{(\mathbf{y}_{\mathrm{exp}})}$ denote their respective attributions. Self-consistency is measured by an alignment function
\[
\alpha\!\left(\boldsymbol{\phi}^{(\mathbf{y}_{\mathrm{dec}})}, \boldsymbol{\phi}^{(\mathbf{y}_{\mathrm{exp}})}\right) : \mathbb{R}^m \times \mathbb{R}^m \rightarrow \mathbb{R},
\]
where the choice of $\alpha$ determines the aspect of the alignment to be evaluated (e.g. magnitude, rank, or overlap). Specific metrics are detailed in the following subsections.

\subsection{Constructing Attribution-Based Preference Data}
\label{sec:build_dataset}

Building on attribution vectors (Section~\ref{sec:seq_attr}) and alignment measures (Section~\ref{sec:att_alig}), we construct a dataset that pairs model outputs with their attributions, allowing systematic comparison between a model’s decision rationale and its post-hoc justifications. Each instance is represented as a quintuple
\[
\big(\mathbf{x},\; \mathbf{y}_{\mathrm{dec}},\; \{ \mathbf{y}_{\mathrm{exp}}^{(i)} \}_{i=1}^{k},\; \boldsymbol{\phi}^{(\mathbf{y}_{\mathrm{dec}})},\; \{ \boldsymbol{\phi}^{(\mathbf{y}_{\mathrm{exp}}^{(i)})} \}_{i=1}^{k} \big),
\]
where $\mathbf{x}$ is the input, $\mathbf{y}_{\mathrm{dec}}$ the predicted answer, and $\{ \mathbf{y}_{\mathrm{exp}}^{(i)} \}_{i=1}^k$ are diverse explanations sampled in a zero-shot setting by conditioning on $(\mathbf{x}, \mathbf{y}_{\mathrm{dec}})$. The corresponding attribution vectors $\boldsymbol{\phi}$ are computed as in Section~\ref{sec:seq_attr}, with non-semantic tokens excluded. We then score each explanation by its alignment $\alpha$ with the decision (Section~\ref{sec:att_alig}), and retain the highest- and lowest-scoring cases as $\mathbf{y}_{\mathrm{exp}}^{\text{chosen}}$ and $\mathbf{y}_{\mathrm{exp}}^{\text{rejected}}$. These preference pairs form the core of the dataset used for optimization (Section~\ref{sec:pref_opt}).

\subsection{Experimental Setup}
\label{sec:exp_setup}

\paragraph{Models.}
We evaluate two instruction-tuned LLaMA models, LLaMA-3.1-8B-Instruct and LLaMA-3.2-3B-Instruct~\cite{touvron2023llama}, in a zero-shot setting, using both the original Meta release and the \textsc{Unsloth}-optimized implementation (identical weights, minor decoding differences).

\paragraph{Datasets.}
Experiments cover four multiple-choice QA datasets: ECQA~\cite{aggarwal2021explanations}, ARC-Easy, ARC-Challenge~\cite{clark2018think}, and CODAH~\cite{chen-etal-2019-codah}, spanning diverse reasoning domains. Each dataset is converted into an attribution-enhanced format (Section~\ref{sec:build_dataset}) and split into train/validation/test (70\%/20\%/10\%) with a fixed seed. Training instances are used to sample $k=5$ explanations and construct preference pairs, validation is used for model selection, and testing for final evaluation.

\paragraph{Prompting and Generation.}
Decisions are elicited using a minimal task instruction (e.g., ``Choose the most plausible answer:''), and explanations are generated as in Section~\ref{sec:build_dataset}, conditioned on $(\mathbf{x}, \mathbf{y}_{\mathrm{dec}})$. Explanations are sampled with nucleus decoding ($p=0.9$, $T=0.7$) and a maximum length of 400 tokens. The complete prompt templates are provided in Appendix~\ref{apx:prompt-templates}.

\paragraph{Feature Attribution.}
We computed attributions using Local Interpretable Model-Agnostic Explanations (LIME)~\cite{ribeiro2016should} and Layer Integrated Gradients (LIG)~\cite{sundararajan2017axiomatic} as implemented in Captum~\cite{Kokhlikyan2020CaptumAU}, with sequence log-probability as the target. 
LIME used 500 perturbation samples with the padding token as baseline, and LIG used an embedding baseline with 25 interpolation steps. 
We applied the \textsc{Unsloth} variants for LIME computations and the original Meta models for LIG to ensure compatibility. 
Skip tokens were filtered as listed in Appendix~\ref{apx:skip_tokens_list}.

\paragraph{Consistency Metrics.} 
We evaluated self-consistency with two alignment functions. \emph{Cosine similarity} measures directional overlap between the attribution vectors, assessing whether explanations emphasize the same features with a similar magnitude. \emph{Spearman rank correlation} captures agreement in feature prioritization while abstracting from the attribution scale:
\[
\mathrm{CC}_{\mathrm{sp}} = 1 - \frac{6 \sum_{i=1}^m 
\left( \mathrm{r}\!\left( \phi^{(\mathbf{y}_{\mathrm{dec}})}_i \right) - 
       \mathrm{r}\!\left( \phi^{(\mathbf{y}_{\mathrm{exp}})}_i \right) \right)^2}
{m(m^2 - 1)},
\]

\noindent where $\mathrm{r}(\cdot)$ assigns the ordinal ranks (ties are decided by average). High values indicate that explanations highlight features aligned with the decision.

\begin{figure}[htbp]
    \centering
    \includegraphics[width=\linewidth]{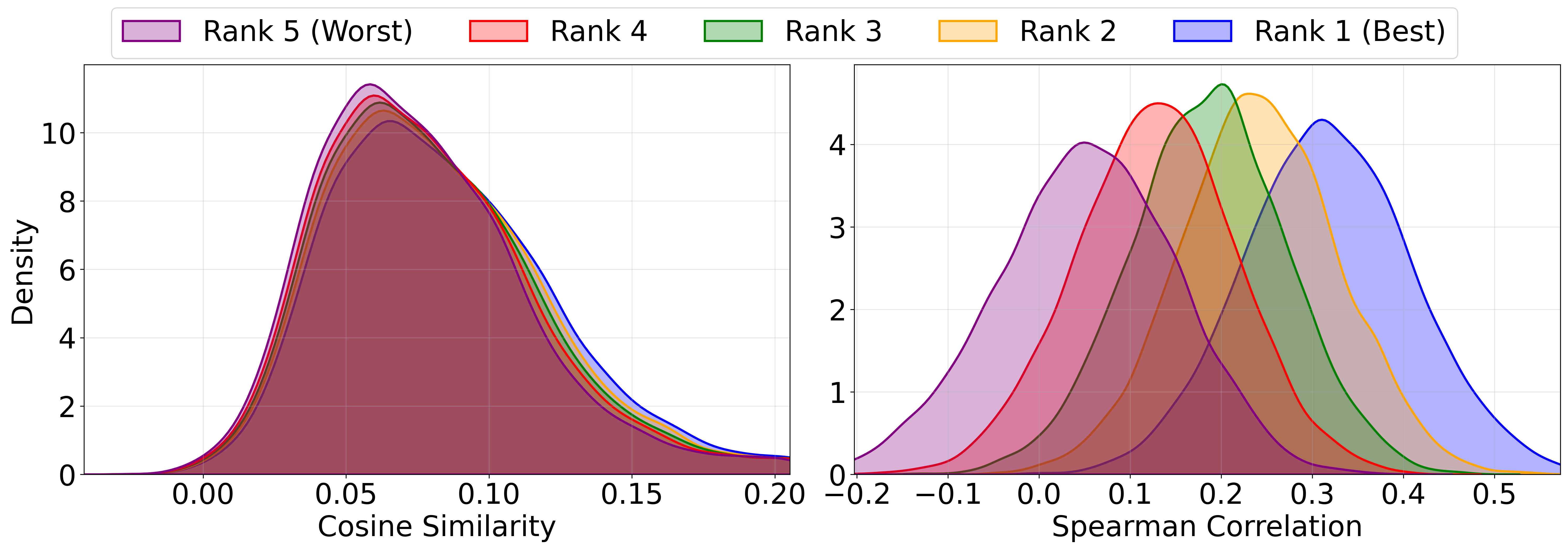}
    \caption{Smoothed distribution of self-consistency scores by explanation rank for the LLaMA3.1-8B model on ECQA. \textbf{Spearman correlation (right)} shows clear rank separation, reflecting strong sensitivity to explanation quality. \textbf{Cosine similarity (left)} shows substantial overlap across ranks, indicating weaker differentiation.}
    \label{fig:spearman_vs_cosine_ranks}
\end{figure}

\begin{figure*}[htbp]
    \centering
    \includegraphics[width=0.9\linewidth]{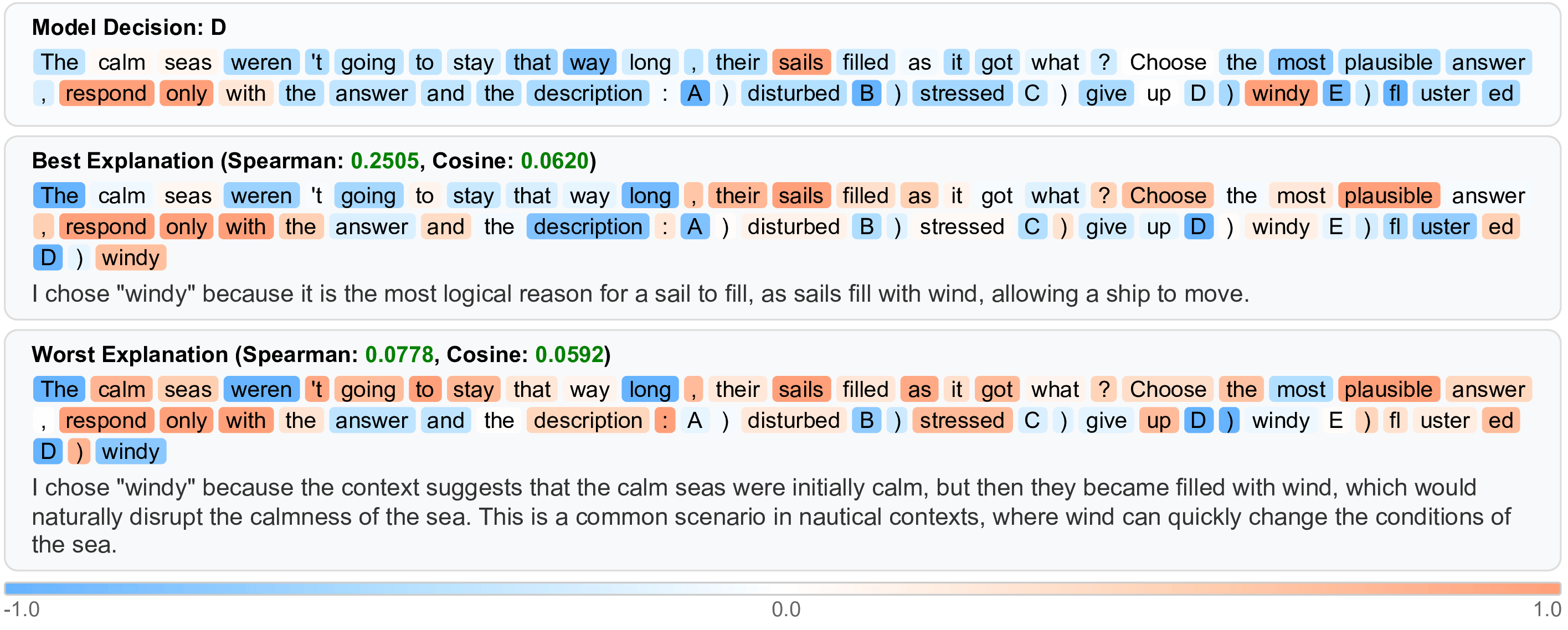}
    \caption{An example from the ECQA dataset shows attribution alignment for the LLaMA3.2-3B model’s decision and its best ($Sp = 0.25$) and worst ($Sp = 0.08$) explanations. The model selects answer “D” (“windy”), with attribution heatmaps highlighting word-level influences (\textcolor{blue}{blue} = negative, \textcolor{orange}{orange} = positive). Key tokens like “sails” and “windy” drive the decision. The best explanation mirrors this focus, correctly reasoning that wind fills sails, while the worst shifts emphasis to “calm seas,” misaligning with the core mechanism.}
    \label{fig:example2}
\end{figure*}

\subsection{Key Findings and Insights}
\label{sec:pcb_res}

We report self-consistency across all model–dataset pairs using both LIME and LIG attributions, measured by cosine similarity (\textsc{CC-Cos}) and Spearman rank correlation (\textsc{CC-Sp}). For each setting, we report the \textit{worst}, \textit{mean}, and \textit{best} scores across five sampled explanations.

\begin{table}[htb]
\small
\centering
\begin{tabularx}{\columnwidth}{l l l | Y Y | Y Y}
\toprule
 & & & \multicolumn{2}{c}{\textbf{LLaMA3.1-8B}} & \multicolumn{2}{c}{\textbf{LLaMA3.2-3B}} \\
\cmidrule(lr){4-5} \cmidrule(lr){6-7}
 &                  & & LIME & LIG & LIME & LIG \\
\specialrule{1pt}{1pt}{3pt}

\multirow{7}{*}{\rotatebox[origin=c]{90}{\textbf{ECQA} (\#10882)}}
  & Acc.       &      & 71.11 & 66.51 & 65.85 & 64.83 \\
  \cmidrule(lr){2-7}
  & \multirow{3}{*}{\rotatebox[origin=c]{90}{CC-Cos}}
                      & Worst & 07.70 & 01.36 & 01.30 & 01.82 \\
  &                   & Mean  & 08.18 & 01.67 & 01.65 & 02.25 \\
  &                   & Best  & 08.67 & 01.98 & 01.99 & 02.66 \\
  \cmidrule(lr){2-7}
  & \multirow{3}{*}{\rotatebox[origin=c]{90}{CC-Sp}}
                      & Worst & 05.07 & 33.79 & 09.75 & 30.55 \\
  &                   & Mean  & 18.47 & 41.70 & 22.28 & 38.72 \\
  &                   & Best  & 31.76 & 49.11 & 34.64 & 46.54 \\

\specialrule{1pt}{1pt}{3pt}

\multirow{7}{*}{\rotatebox[origin=c]{90}{\textbf{ARC-E} (\#5197)}}
  & Acc.       &      & 87.01 & 82.87 & 81.90 & 80.69 \\
  \cmidrule(lr){2-7}
  & \multirow{3}{*}{\rotatebox[origin=c]{90}{CC-Cos}}
                      & Worst & 13.31 & 01.49 & 05.32 & 01.65 \\
  &                   & Mean  & 14.16 & 01.78 & 05.86 & 02.04 \\
  &                   & Best  & 15.02 & 02.07 & 06.41 & 02.41 \\
  \cmidrule(lr){2-7}
  & \multirow{3}{*}{\rotatebox[origin=c]{90}{CC-Sp}}
                      & Worst & \textminus01.06 & 38.81 & 07.26 & 36.63 \\
  &                   & Mean  & 12.44 & 46.42 & 19.68 & 45.58 \\
  &                   & Best  & 25.77 & 53.41 & 31.98 & 53.68 \\

\specialrule{1pt}{1pt}{3pt}

\multirow{7}{*}{\rotatebox[origin=c]{90}{\textbf{ARC-C} (\#2590)}}
  & Acc.       &      & 77.57 & 71.12 & 68.70 & 64.90 \\
  \cmidrule(lr){2-7}
  & \multirow{3}{*}{\rotatebox[origin=c]{90}{CC-Cos}}
                      & Worst & 15.09 & 01.39 & 07.67 & 01.49 \\
  &                   & Mean  & 16.07 & 01.67 & 08.29 & 01.90 \\
  &                   & Best  & 17.08 & 01.94 & 08.93 & 02.29 \\
  \cmidrule(lr){2-7}
  & \multirow{3}{*}{\rotatebox[origin=c]{90}{CC-Sp}}
                      & Worst & \textminus01.82 & 37.07 & 05.94 & 33.22 \\
  &                   & Mean  & 11.48 & 44.87 & 18.44 & 41.43 \\
  &                   & Best  & 24.73 & 52.11 & 30.89 & 49.12 \\

\specialrule{1pt}{1pt}{3pt}

\multirow{7}{*}{\rotatebox[origin=c]{90}{\textbf{CODAH} (\#2776)}}
  & Acc.       &      & 83.39 & -- & 75.48 & 72.05 \\
  \cmidrule(lr){2-7}
  & \multirow{3}{*}{\rotatebox[origin=c]{90}{CC-Cos}}
                      & Worst & 12.87 & -- & 07.49 & 01.91 \\
  &                   & Mean  & 13.68 & -- & 08.06 & 02.47 \\
  &                   & Best  & 14.52 & -- & 08.64 & 02.96 \\
  \cmidrule(lr){2-7}
  & \multirow{3}{*}{\rotatebox[origin=c]{90}{CC-Sp}}
                      & Worst & 07.11 & -- & 06.82 & 35.24 \\
  &                   & Mean  & 19.97 & -- & 19.39 & 44.88 \\
  &                   & Best  & 32.66 & -- & 31.95 & 53.60 \\

\bottomrule
\end{tabularx}
\caption{PSCB self-consistency scores ($\times 100$) on the test split of each dataset, for both LLaMA models and both attribution methods (LIME, LIG). Worst, Mean, and Best are reported over 5 sampled explanations per item. CC-Cos: cosine similarity; CC-Sp: Spearman rank correlation between decision and explanation attribution vectors.}
\label{tab:post-hoc_consistency_bank}
\end{table}
\begin{table}[htbp]
\centering
\small
\setlength{\tabcolsep}{3.5pt}
\renewcommand{\arraystretch}{1.3}
\begin{tabular}{llcc}
\toprule
 & \textbf{Model} & \textbf{CC-Cos (T / F / $\Delta$)} & \textbf{CC-Sp (T / F / $\Delta$)} \\
\midrule
\multirow{2}{*}{\rotatebox[origin=c]{90}{\scriptsize\textbf{ECQA}}}
  & LLaMA3.1-8B & 8.1 / 8.4 / \textbf{\textminus0.3} & 18.6 / 18.2 / \textbf{+0.4} \\
  & LLaMA3.2-3B & 1.6 / 1.7 / \textbf{\textminus0.1} & 22.3 / 22.3 / \textbf{0.0} \\
\midrule
\multirow{2}{*}{\rotatebox[origin=c]{90}{\scriptsize\textbf{ARC-E}}}
  & LLaMA3.1-8B & 14.1 / 14.8 / \textbf{\textminus0.7} & 12.4 / 13.0 / \textbf{\textminus0.6} \\
  & LLaMA3.2-3B & 5.7 / 6.6 / \textbf{\textminus0.9} & 19.8 / 19.3 / \textbf{+0.5} \\
\midrule
\multirow{2}{*}{\rotatebox[origin=c]{90}{\scriptsize\textbf{ARC-C}}}
  & LLaMA3.1-8B & 16.1 / 16.1 / \textbf{0.0} & 11.5 / 11.5 / \textbf{0.0} \\
  & LLaMA3.2-3B & 8.6 / 7.6 / \textbf{+1.0} & 18.5 / 18.4 / \textbf{+0.1} \\
\midrule
\multirow{2}{*}{\rotatebox[origin=c]{90}{\scriptsize\textbf{CODAH}}}
  & LLaMA3.1-8B & 13.7 / 13.4 / \textbf{+0.3} & 19.8 / 20.9 / \textbf{\textminus1.1} \\
  & LLaMA3.2-3B & 8.2 / 7.5 / \textbf{+0.7} & 19.2 / 19.9 / \textbf{\textminus0.7} \\
\bottomrule
\end{tabular}
\caption{
Self-consistency scores (×100) for true (T) and false (F) predictions. We denote $\Delta = T - F.$  Cosine similarity and Spearman rank correlation are shown. Differences are small and vary in direction.
}
\label{tab:correctness_consistency_scores}
\end{table}

\paragraph{Explanation variability and metric differences.}
We observe substantial gaps between the best and worst explanations for a given input. For example, in ECQA with LLaMA-3.1-8B (LIME attributions), Spearman scores range from 5.07 to 31.76. Across datasets and models, Spearman correlations are consistently higher than cosine similarities, indicating that models preserve feature \textit{ranking} more reliably than attribution magnitudes. Spearman also exhibits greater spread across explanations, making it more effective in detecting consistency differences, whereas cosine tends to produce compressed values due to its sensitivity to attribution scale. This supports the need for ranking-based evaluation and highlights the opportunity to identify highly self-consistent explanations among diverse generations (Figure~\ref{fig:spearman_vs_cosine_ranks}). Figure~\ref{fig:example2} further shows that a large Spearman variance corresponds to meaningful semantic differences in explanation quality.

\paragraph{Correctness and self-consistency.}
We test whether self-consistency correlates with prediction correctness by comparing alignment scores for correct vs.\ incorrect predictions (Table~\ref{tab:correctness_consistency_scores} in Appendix). Differences are small and inconsistent across datasets and models, suggesting that self-consistency is largely orthogonal to accuracy.

\section{Improving Self-Consistency with DPO}
\label{sec:pref_opt}
\newcolumntype{Z}{>{\hsize=0.9\hsize\centering\arraybackslash}X}  

\newcolumntype{T}{>{\centering\arraybackslash}p{2.6cm}}  
\newcolumntype{S}{>{\centering\arraybackslash}p{1.6cm}}  
\newcolumntype{L}[1]{>{\RaggedRight\arraybackslash}p{#1}}
\setlength{\tabcolsep}{1.45pt}  

\begin{table*}[tbh]
\small
\centering
\begin{tabularx}{\textwidth}{l l l l @{\hskip 2pt\vrule width 0.4pt\hskip 2pt} S T T @{\hskip 2pt\vrule width 0.4pt\hskip 2pt} S T T}
\toprule
 & & & &
 \multicolumn{3}{c}{\textbf{LLaMA3.1-8B}} &
 \multicolumn{3}{c}{\textbf{LLaMA3.2-3B}} \\
 & & & & B-B & B-T~\scriptsize($\uparrow\downarrow$) & T-T~\scriptsize($\uparrow\downarrow$) & B-B & B-T~\scriptsize($\uparrow\downarrow$) & T-T~\scriptsize($\uparrow\downarrow$) \\

\specialrule{1pt}{1pt}{3pt}

\multirow{14}{*}{\rotatebox[origin=c]{90}{\textbf{ECQA}}}
  & \multirow{7}{*}{\rotatebox[origin=c]{90}{LIME}}
  & & Acc. & 70.34 & 70.34 & 69.70~\downDelta{0.9\%} & 67.13 & 67.13 & 66.12~\downDelta{1.5\%} \\
  \cmidrule(lr){3-10}
  & & \multirow{3}{*}{\rotatebox[origin=c]{90}{CC-Cos}}
      & Worst & 07.58 {\scriptsize$\pm 0.11$} & 07.64 {\scriptsize$\pm 0.12$}~\upDelta{0.8\%} & \textbf{09.16} {\scriptsize$\pm 0.12$}~\upDelta{20.8\%} & 01.11 {\scriptsize$\pm 0.14$} & 01.30 {\scriptsize$\pm 0.13$}~\upDelta{17.1\%} & \textbf{02.14} {\scriptsize$\pm 0.13$}~\upDelta{92.8\%} \\
  & &  & Mean  & 08.05 {\scriptsize$\pm 0.12$} & 08.10 {\scriptsize$\pm 0.12$}~\upDelta{0.6\%} & \textbf{09.72} {\scriptsize$\pm 0.12$}~\upDelta{20.7\%} & 01.46 {\scriptsize$\pm 0.14$} & 01.63 {\scriptsize$\pm 0.13$}~\upDelta{11.6\%} & \textbf{02.47} {\scriptsize$\pm 0.14$}~\upDelta{69.2\%} \\
  & &  & Best  & 08.53 {\scriptsize$\pm 0.13$} & 08.57 {\scriptsize$\pm 0.13$}~\upDelta{0.5\%} & \textbf{10.28} {\scriptsize$\pm 0.13$}~\upDelta{20.5\%} & 01.81 {\scriptsize$\pm 0.14$} & 01.96 {\scriptsize$\pm 0.13$}~\upDelta{8.3\%} & \textbf{02.80} {\scriptsize$\pm 0.14$}~\upDelta{54.7\%} \\
  \cmidrule(lr){3-10}
  & & \multirow{3}{*}{\rotatebox[origin=c]{90}{CC-Sp}}
      & Worst & 05.00 {\scriptsize$\pm 0.30$} & 07.20 {\scriptsize$\pm 0.30$}~\upDelta{44.0\%} & \textbf{07.86} {\scriptsize$\pm 0.29$}~\upDelta{57.2\%} & 10.04 {\scriptsize$\pm 0.31$} & 10.50 {\scriptsize$\pm 0.32$}~\upDelta{4.6\%} & \textbf{12.94} {\scriptsize$\pm 0.32$}~\upDelta{28.9\%} \\
  & &  & Mean  & 18.22 {\scriptsize$\pm 0.23$} & 20.22 {\scriptsize$\pm 0.23$}~\upDelta{11.0\%} & \textbf{20.58} {\scriptsize$\pm 0.24$}~\upDelta{13.0\%} & 22.52 {\scriptsize$\pm 0.26$} & 22.65 {\scriptsize$\pm 0.27$}~\upDelta{0.6\%} & \textbf{24.71} {\scriptsize$\pm 0.27$}~\upDelta{9.7\%} \\
  & &  & Best  & 31.30 {\scriptsize$\pm 0.28$} & 32.86 {\scriptsize$\pm 0.29$}~\upDelta{5.0\%} & \textbf{33.45} {\scriptsize$\pm 0.28$}~\upDelta{6.9\%} & 35.02 {\scriptsize$\pm 0.30$} & 34.81 {\scriptsize$\pm 0.30$}~\downDelta{0.6\%} & \textbf{36.50} {\scriptsize$\pm 0.29$}~\upDelta{4.2\%} \\

  \cmidrule(lr){2-10}
  
  & \multirow{7}{*}{\rotatebox[origin=c]{90}{LIG}}
  & & Acc. & 67.03 & 67.03 & 67.68~\upDelta{1.1\%} & 62.44 & 62.44 & 62.99~\upDelta{0.9\%} \\
  \cmidrule(lr){3-10}
  
  & & \multirow{3}{*}{\rotatebox[origin=c]{90}{CC-Cos}}
      & Worst & 0.87 {\scriptsize$\pm$ 0.03} & 01.35 {\scriptsize$\pm$ 0.02}~\upDelta{55.2\%} & \textbf{01.36} {\scriptsize$\pm$ 0.02}~\upDelta{56.3\%} & 01.90 {\scriptsize$\pm$ 0.03} & 01.97 {\scriptsize$\pm$ 0.03}~\upDelta{3.7\%} & \textbf{02.54} {\scriptsize$\pm$ 0.03}~\upDelta{33.7\%} \\
  & &  & Mean  & 01.22 {\scriptsize$\pm$ 0.03} & 01.67 {\scriptsize$\pm$ 0.02}~\upDelta{36.9\%} & \textbf{01.67} {\scriptsize$\pm$ 0.02}~\upDelta{36.9\%} & 02.32 {\scriptsize$\pm$ 0.03} & 02.37 {\scriptsize$\pm$ 0.03}~\upDelta{2.2\%} & \textbf{02.82} {\scriptsize$\pm$ 0.03}~\upDelta{21.6\%} \\
  & &  & Best  & 01.55 {\scriptsize$\pm$ 0.03} & 01.96 {\scriptsize$\pm$ 0.03}~\upDelta{26.5\%} & \textbf{01.96} {\scriptsize$\pm$ 0.02}~\upDelta{26.5\%} & 02.72 {\scriptsize$\pm$ 0.03} & 02.73 {\scriptsize$\pm$ 0.04}~\upDelta{0.4\%} & \textbf{02.99} {\scriptsize$\pm$ 0.03}~\upDelta{9.9\%} \\
  \cmidrule(lr){3-10}
  
  & & \multirow{3}{*}{\rotatebox[origin=c]{90}{CC-Sp}}
      & Worst & 34.71 {\scriptsize$\pm$ 0.44} & 34.71 {\scriptsize$\pm$ 0.45} & \textbf{36.59} {\scriptsize$\pm$ 0.42}~\upDelta{5.4\%} & 28.61 {\scriptsize$\pm$ 0.45} & 29.04 {\scriptsize$\pm$ 0.45}~\upDelta{1.5\%} & \textbf{33.30} {\scriptsize$\pm$ 0.45}~\upDelta{16.4\%} \\
  & &  & Mean  & 42.52 {\scriptsize$\pm$ 0.40} & 42.65 {\scriptsize$\pm$ 0.41}~\upDelta{0.3\%} & \textbf{43.66} {\scriptsize$\pm$ 0.39}~\upDelta{2.7\%} & 36.72 {\scriptsize$\pm$ 0.42} & 37.17 {\scriptsize$\pm$ 0.41}~\upDelta{1.2\%} & \textbf{41.90} {\scriptsize$\pm$ 0.41}~\upDelta{14.1\%} \\
  & &  & Best  & 49.87 {\scriptsize$\pm$ 0.40} & 50.15 {\scriptsize$\pm$ 0.40}~\upDelta{0.6\%} & \textbf{51.24} {\scriptsize$\pm$ 0.38}~\upDelta{2.7\%} & 44.53 {\scriptsize$\pm$ 0.41} & 45.05 {\scriptsize$\pm$ 0.41}~\upDelta{1.2\%} & \textbf{50.17} {\scriptsize$\pm$ 0.41}~\upDelta{12.7\%} \\

\specialrule{1pt}{1pt}{3pt}

\multirow{14}{*}{\rotatebox[origin=c]{90}{\textbf{ARC-Easy}}}
  & \multirow{7}{*}{\rotatebox[origin=c]{90}{LIME}} & & Acc. & 87.72 & 87.72 & 86.76~\downDelta{1.1\%} & 81.62 & 81.62 & 81.04~\downDelta{0.7\%} \\
  \cmidrule(lr){3-10}
  & & \multirow{3}{*}{\rotatebox[origin=c]{90}{CC-Cos}}
      & Worst & 13.32 {\scriptsize$\pm 0.25$} & 13.47 {\scriptsize$\pm 0.26$}~\upDelta{1.1\%} & \textbf{14.97} {\scriptsize$\pm 0.24$}~\upDelta{12.4\%} & 05.58 {\scriptsize$\pm 0.32$} & 05.58 {\scriptsize$\pm 0.30$} & \textbf{06.32} {\scriptsize$\pm 0.29$}~\upDelta{13.2\%} \\
  & &  & Mean  & 14.17 {\scriptsize$\pm 0.27$} & 14.31 {\scriptsize$\pm 0.27$}~\upDelta{1.0\%} & \textbf{15.89} {\scriptsize$\pm 0.25$}~\upDelta{12.1\%} & 06.13 {\scriptsize$\pm 0.33$} & 06.11 {\scriptsize$\pm 0.31$}~\downDelta{0.3\%} & \textbf{06.87} {\scriptsize$\pm 0.30$}~\upDelta{12.1\%} \\
  & &  & Best  & 15.03 {\scriptsize$\pm 0.28$} & 15.15 {\scriptsize$\pm 0.29$}~\upDelta{0.8\%} & \textbf{16.82} {\scriptsize$\pm 0.27$}~\upDelta{11.9\%} & 06.68 {\scriptsize$\pm 0.34$} & 06.66 {\scriptsize$\pm 0.32$}~\downDelta{2.9\%} & \textbf{07.42} {\scriptsize$\pm 0.31$}~\upDelta{11.1\%} \\
  \cmidrule(lr){3-10}
  & & \multirow{3}{*}{\rotatebox[origin=c]{90}{CC-Sp}}
      & Worst & -01.90 {\scriptsize$\pm 0.43$} & \textbf{01.84} {\scriptsize$\pm 0.44$}~\upDelta{96.8\%} & 01.76 {\scriptsize$\pm 0.44$}~\upDelta{92.6\%} & 07.97 {\scriptsize$\pm 0.45$} & 08.99 {\scriptsize$\pm 0.44$}~\upDelta{12.8\%} & \textbf{11.48} {\scriptsize$\pm 0.46$}~\upDelta{44.0\%} \\
  & &  & Mean  & 12.01 {\scriptsize$\pm 0.35$} & \textbf{15.01} {\scriptsize$\pm 0.38$}~\upDelta{25.0\%} & 14.88 {\scriptsize$\pm 0.37$}~\upDelta{23.9\%} & 20.72 {\scriptsize$\pm 0.39$} & 21.53 {\scriptsize$\pm 0.39$}~\upDelta{3.9\%} & \textbf{23.16} {\scriptsize$\pm 0.39$}~\upDelta{11.8\%} \\
  & &  & Best  & 25.72 {\scriptsize$\pm 0.43$} & \textbf{28.38} {\scriptsize$\pm 0.45$}~\upDelta{10.3\%} & 27.86 {\scriptsize$\pm 0.44$}~\upDelta{8.3\%} & 33.14 {\scriptsize$\pm 0.44$} & 34.19 {\scriptsize$\pm 0.46$}~\upDelta{3.2\%} & \textbf{34.80} {\scriptsize$\pm 0.41$}~\upDelta{5.0\%} \\
 
  \cmidrule(lr){2-10}
  
  & \multirow{7}{*}{\rotatebox[origin=c]{90}{LIG}}
  & & Acc. & 84.26 & 84.26 & 84.84~\upDelta{0.7\%} & 80.23 & 80.23 & 79.27~\downDelta{1.2\%} \\
  \cmidrule(lr){3-10}
  & & \multirow{3}{*}{\rotatebox[origin=c]{90}{CC-Cos}}
      & Worst & 0.98 {\scriptsize$\pm$ 0.05} & \textbf{01.56} {\scriptsize$\pm$ 0.04}~\upDelta{59.2\%} & 01.54 {\scriptsize$\pm$ 0.04}~\upDelta{57.1\%} & 01.20 {\scriptsize$\pm$ 0.04} & 01.25 {\scriptsize$\pm$ 0.04}~\upDelta{4.2\%} & \textbf{01.71} {\scriptsize$\pm$ 0.04}~\upDelta{42.5\%} \\
  & &  & Mean  & 01.35 {\scriptsize$\pm$ 0.05} & \textbf{01.84} {\scriptsize$\pm$ 0.04}~\upDelta{36.3\%} & 01.82 {\scriptsize$\pm$ 0.04}~\upDelta{34.8\%} & 01.62 {\scriptsize$\pm$ 0.04} & 01.66 {\scriptsize$\pm$ 0.04}~\upDelta{2.5\%} & \textbf{02.07} {\scriptsize$\pm$ 0.04}~\upDelta{27.8\%} \\
  & &  & Best  & 01.68 {\scriptsize$\pm$ 0.05} & \textbf{02.11} {\scriptsize$\pm$ 0.04}~\upDelta{25.6\%} & 02.08 {\scriptsize$\pm$ 0.04}~\upDelta{23.8\%} & 02.01 {\scriptsize$\pm$ 0.04} & 02.03 {\scriptsize$\pm$ 0.04}~\upDelta{1.0\%} & \textbf{02.43} {\scriptsize$\pm$ 0.04}~\upDelta{20.9\%} \\
  \cmidrule(lr){3-10}
  & & \multirow{3}{*}{\rotatebox[origin=c]{90}{CC-Sp}}
      & Worst & 37.40 {\scriptsize$\pm$ 0.64} & 38.94 {\scriptsize$\pm$ 0.66}~\upDelta{4.1\%} & \textbf{39.93} {\scriptsize$\pm$ 0.65}~\upDelta{6.8\%} & 33.77 {\scriptsize$\pm$ 0.65} & 34.88 {\scriptsize$\pm$ 0.67}~\upDelta{3.3\%} & \textbf{37.97} {\scriptsize$\pm$ 0.65}~\upDelta{12.4\%} \\
  & &  & Mean  & 45.55 {\scriptsize$\pm$ 0.55} & 46.76 {\scriptsize$\pm$ 0.57}~\upDelta{2.7\%} & \textbf{47.09} {\scriptsize$\pm$ 0.58}~\upDelta{3.4\%} & 43.15 {\scriptsize$\pm$ 0.60} & 44.02 {\scriptsize$\pm$ 0.61}~\upDelta{2.0\%} & \textbf{46.24} {\scriptsize$\pm$ 0.60}~\upDelta{7.2\%} \\
  & &  & Best  & 52.88 {\scriptsize$\pm$ 0.52} & 53.63 {\scriptsize$\pm$ 0.53}~\upDelta{1.4\%} & \textbf{53.79} {\scriptsize$\pm$ 0.55}~\upDelta{1.7\%} & 51.75 {\scriptsize$\pm$ 0.60} & 52.52 {\scriptsize$\pm$ 0.60}~\upDelta{1.5\%} & \textbf{53.74} {\scriptsize$\pm$ 0.60}~\upDelta{3.8\%} \\

\bottomrule
\end{tabularx}
\caption{In-Domain Performance: Models trained and evaluated on the same dataset (ECQA, ARC-Easy). Results are based on LIME and LIG attributions, reporting Worst, Mean, and Best self-consistency scores across 5 sampled explanations (×100). Accuracy is shown alongside. Each model is evaluated under three modes (B--B, B--T, T--T), with relative improvements over B--B indicated by arrows.}
\label{tab:in_domain_eval}
\end{table*}
\begin{table*}[bht]
\centering
\small
\begin{tabularx}{\textwidth}{l l l @{\hskip 2pt\vrule width 0.4pt\hskip 2pt} >{\centering\arraybackslash}p{1cm} @{\hskip 2pt\vrule width 0.4pt\hskip 2pt} cc @{\hskip 2pt\vrule width 0.4pt\hskip 2pt} cc}
\toprule
& & \textbf{Training Data} &
\textbf{Acc.} &
\multicolumn{2}{c}{\textbf{CC-Cos~\scriptsize($\uparrow\downarrow$)}} &
\multicolumn{2}{c}{\textbf{CC-Sp~\scriptsize($\uparrow\downarrow$)}} \\
\cmidrule(lr){5-6} \cmidrule(lr){7-8}
 & & & & B-T & T-T & B-T & T-T \\
\midrule
\multirow{6}{*}{\rotatebox[origin=c]{90}{\textbf{CODAH}}} & \multirow{3}{*}{L3.1-8B} 
& None & 81.29 & \multicolumn{2}{c@{\hskip 2pt\vrule width 0.4pt\hskip 2pt}}{13.50 ± 0.28} & \multicolumn{2}{c}{19.46 ± 0.48} \\
&& ECQA & 77.70 & 13.39 ± 0.27~\scriptsize(\textcolor{firebrick}{$\downarrow$ 0.8\%}) & 14.05 ± 0.32~\scriptsize(\textcolor{seagreen}{$\uparrow$ 4.1\%}) & 18.30 ± 0.56~\scriptsize(\textcolor{firebrick}{$\downarrow$ 6.0\%}) & 21.58 ± 0.49~\scriptsize(\textcolor{seagreen}{$\uparrow$ 10.9\%}) \\
&& ARC-Easy & 80.94 & 13.38 ± 0.27~\scriptsize(\textcolor{firebrick}{$\downarrow$ 0.8\%}) & \textbf{14.29} ± 0.29~\scriptsize(\textcolor{seagreen}{$\uparrow$ 5.9\%}) & 18.79 ± 0.57~\scriptsize(\textcolor{firebrick}{$\downarrow$ 3.4\%}) & \textbf{21.71} ± 0.50~\scriptsize(\textcolor{seagreen}{$\uparrow$ 11.6\%}) \\
\cmidrule{2-8}
& \multirow{3}{*}{L3.2-3B}
& None & 73.90 & \multicolumn{2}{c@{\hskip 2pt\vrule width 0.4pt\hskip 2pt}}{08.04 ± 0.34} & \multicolumn{2}{c}{17.95 ± 0.57} \\
&& ECQA & 72.43 & 08.16 ± 0.35~\scriptsize(\textcolor{seagreen}{$\uparrow$ 1.5\%}) & 09.42 ± 0.31~\scriptsize(\textcolor{seagreen}{$\uparrow$ 17.2\%}) & 19.41 ± 0.54~\scriptsize(\textcolor{seagreen}{$\uparrow$ 8.1\%}) & 20.10 ± 0.55~\scriptsize(\textcolor{seagreen}{$\uparrow$ 12.0\%}) \\
&& ARC-Easy & 74.63 & 08.02 ± 0.34 ~\scriptsize(\textcolor{firebrick}{$\downarrow$ 0.2\%}) & \textbf{09.51} ± 0.31~\scriptsize(\textcolor{seagreen}{$\uparrow$ 18.3\%}) & 18.13 ± 0.57~\scriptsize(\textcolor{seagreen}{$\uparrow$ 1.0\%}) & \textbf{20.52} ± 0.55~\scriptsize(\textcolor{seagreen}{$\uparrow$ 14.3\%}) \\

\midrule

\multirow{6}{*}{\rotatebox[origin=c]{90}{\textbf{ARC-Chal}}} & \multirow{3}{*}{L3.1-8B} 
& None & 76.15 & \multicolumn{2}{c@{\hskip 2pt\vrule width 0.4pt\hskip 2pt}}{16.26 ± 0.33} & \multicolumn{2}{c}{11.81 ± 0.45} \\
&& ECQA & 73.85 & 16.48 ± 0.33~\scriptsize(\textcolor{seagreen}{$\uparrow$ 1.3\%}) & 17.57 ± 0.32~\scriptsize(\textcolor{seagreen}{$\uparrow$ 8.1\%}) & 13.60 ± 0.49~\scriptsize(\textcolor{seagreen}{$\uparrow$ 13.16\%}) & 13.45 ± 0.49~\scriptsize(\textcolor{seagreen}{$\uparrow$ 13.9\%}) \\
&& ARC-Easy & 73.46 & 16.49 ± 0.33~\scriptsize(\textcolor{seagreen}{$\uparrow$ 1.4\%}) & \textbf{17.65} ± 0.31~\scriptsize(\textcolor{seagreen}{$\uparrow$ 8.5\%}) & \textbf{13.87} ± 0.49~\scriptsize(\textcolor{seagreen}{$\uparrow$ 17.4\%}) & 13.31 ± 0.49~\scriptsize(\textcolor{seagreen}{$\uparrow$ 12.7\%}) \\
\cmidrule{2-8}
& \multirow{3}{*}{L3.2-3B}
& None & 66.67 & \multicolumn{2}{c@{\hskip 2pt\vrule width 0.4pt\hskip 2pt}}{08.58 ± 0.46} & \multicolumn{2}{c}{19.51 ± 0.62} \\
&& ECQA & 66.67 & 08.46 ± 0.44~\scriptsize(\textcolor{firebrick}{$\downarrow$ 1.4\%}) & \textbf{09.13} ± 0.41~\scriptsize(\textcolor{seagreen}{$\uparrow$ 6.4\%}) & 19.83 ± 0.62~\scriptsize(\textcolor{seagreen}{$\uparrow$ 1.6\%}) & 21.37 ± 0.59~\scriptsize(\textcolor{seagreen}{$\uparrow$ 9.5\%}) \\
&& ARC-Easy & 67.06 & 08.49 ± 0.44~\scriptsize(\textcolor{firebrick}{$\downarrow$ 1.0\%}) & 09.08 ± 0.42~\scriptsize(\textcolor{seagreen}{$\uparrow$ 5.8\%}) & 19.83 ± 0.69~\scriptsize(\textcolor{seagreen}{$\uparrow$ 1.6\%}) & \textbf{21.78} ± 0.59~\scriptsize(\textcolor{seagreen}{$\uparrow$ 11.6\%}) \\

\bottomrule
\end{tabularx}
\caption{Cross-Domain Performance: Models trained on source datasets (ECQA, ARC-Easy) and evaluated on target datasets (CODAH, ARC-Challenge). Results are based on LIME evaluations, showing mean Cosine similarity (CC-Cos) and mean Spearman correlation (CC-Sp) (×100), with relative improvements over base models in parentheses.}
\label{tab:cross_domain_eval}
\end{table*}
\begin{table}[t]
\small
\centering
\setlength{\tabcolsep}{2.5pt}
\begin{tabularx}{\columnwidth}{l l | c | Y Y | Y Y}
\toprule
& & \textbf{B-B} & \multicolumn{2}{c|}{\textbf{SFT}} & \multicolumn{2}{c}{\textbf{DPO}} \\
& & & B-T & T-T & B-T & T-T \\
\midrule
& & Acc. & 70.34 & 71.53 & 70.34 & 69.70 \\
\cmidrule{2-7}
\multirow{3}{*}{\rotatebox[origin=c]{90}{CC-Cos}}
& Worst & 07.58 & 07.01 & 07.65 & 07.64 & \textbf{09.16} \\
& Mean  & 08.05 & 07.52 & 08.19 & 08.10 & \textbf{09.72} \\
& Best  & 08.53 & 08.04 & 08.74 & 08.57 & \textbf{10.28} \\
\cmidrule{2-7}
\multirow{3}{*}{\rotatebox[origin=c]{90}{CC-Sp}}
& Worst & 05.00 & -02.66 & -03.46 & 07.20 & \textbf{07.86} \\
& Mean  & 18.22 & 11.42 & 10.68 & 20.22 & \textbf{20.58} \\
& Best  & 31.30 & 25.38 & 24.47 & 32.86 & \textbf{33.45} \\
\bottomrule
\end{tabularx}
\caption{SFT vs.\ DPO on ECQA (LLaMA3.1-8B, LIME). SFT is trained on the highest-ranked explanations (by Spearman correlation) used as the chosen examples in DPO. SFT degrades Spearman correlation across all settings, while DPO consistently improves it.}
\label{tab:sft_vs_dpo}
\end{table}

To directly optimize for self-consistency, we fine-tune language models using Direct Preference Optimization (DPO)~\cite{rafailov2023direct}, which aligns outputs with preference signals through relative comparisons.

\paragraph{Preference Pair Construction.}
For each input $\mathbf{x}$, we construct preference pairs by selecting a preferred explanation $\mathbf{y}^{\text{chosen}}_{\text{exp}}$ and a dispreferred explanation $\mathbf{y}^{\text{rejected}}_{\text{exp}}$ based on their attribution alignment with the model's decision $\mathbf{y}_{\text{dec}}$ (Section~\ref{sec:att_alig}). We rank five candidate explanations using Spearman rank correlation between their attribution vectors and the decision's attribution vector, then select the highest and lowest ranked as the preference pair.

\paragraph{Alignment Metric.}
We employ Spearman rank correlation as our primary alignment metric for DPO training over previously used cosine similarity. Spearman correlation captures feature prioritization rather than magnitude alignment. As demonstrated in Section~\ref{sec:pcb_res} (Figure~\ref{fig:spearman_vs_cosine_ranks}), Spearman correlation exhibits greater score variability across explanation candidates. This broader dynamic range provides a more informative training signal, enabling a clearer differentiation between high-quality and low-quality explanations. 

\paragraph{Training Procedure.}
We use LoRA~\cite{Hu2021LoRALA} for parameter-efficient adaptation, inserting trainable low-rank matrices into the transformer's attention and feed-forward layers. All explanation generations follow the zero-shot prompting scheme of Section~\ref{sec:exp_setup} to ensure train--eval consistency. Models are fine-tuned independently per dataset; hyperparameters and LoRA configurations are in Appendix~\ref{apx:implementation_details}. A qualitative comparison of DPO-tuned and vanilla explanations is shown in Figure~\ref{fig:example3}.

\section{Empirical Evaluation}
\label{sec:empirical_evaluation}
We assess the impact of DPO training across three Decider–Explainer configurations: 
B--B (Base Decider, Base Explainer), B--T (Base Decider, Tuned Explainer), and T--T (Tuned Decider, Tuned Explainer). 
The B--T setting is particularly noteworthy, as it evaluates whether the tuned model can function as a plug-in explainer while keeping the decision model unchanged.

\paragraph{Baseline Definition.}
Improvements are measured relative to the B--B baseline, with all other factors (architecture, prompting, decoding, attribution pipeline) held constant.
The B--B configuration represents a fully untuned baseline, while B--T isolates the effect of tuning the explainer alone.
Comparisons to T--T therefore isolate the effect of attribution-based preference optimization on the decision model.
To our knowledge, no prior methods directly optimize attributional self-consistency. Methods such as PEX~\citep{zhao2025pex} optimize explanation-label probabilistic support, which is complementary but operates at a different level of analysis: a model can achieve strong probabilistic consistency while its explanation still relies on different input features than its decision.

\subsection{In-Domain Evaluation}
\label{sec:in_domain_eval}

We begin with in-domain experiments, where models are fine-tuned and evaluated on the same dataset. 
This setting isolates the direct effect of DPO training on explanation self-consistency, and Table~\ref{tab:in_domain_eval} reports task accuracy alongside self-consistency scores computed with both \textsc{LIME} and \textsc{LIG} in the three Decider-Explainer modes.
Across both model sizes, DPO training consistently improves explanation quality in the
\textbf{T--T mode}, with the strongest gains
for the weakest explanations. On ECQA for LLaMA3.1-8B, Spearman correlation improves by 13\% on mean and 57.2\% on worst-case with LIME, while LIG adds smaller gains of 2.7\% and 5.4\%. On ARC-Easy, LIME nearly doubles the worst-case score (+92.6\%) and raises the mean by 23.9\%, while LIG yields smaller but consistent improvements of 6.8\% and 3.4\%. The smaller LLaMA3.2-3B follows the same trend: on ECQA, LIME boosts mean and worst-case scores by 9.7\% and 28.9\%, while LIG raises them by 14.1\% and 16.4\%; on ARC-Easy, LIME increases the mean by 11.8\% and worst-case by 44.0\%, while LIG improves them by 7.2\% and 12.4\%.
Results in the \textbf{B--T mode} also improve, though more moderately, reflecting the mismatch between a vanilla decider and a tuned explainer. The stronger gains in T--T highlight that self-consistency is highest when the same tuned model generates both decisions and explanations, aligning their attribution vectors.
Across datasets and models, these improvements come with only minimal accuracy changes ($\leq 1.5\%$ in either direction), showing that DPO enhances self-consistency without harming predictive performance.
To verify that these patterns extend beyond the LLaMA family, we apply the same pipeline to \textsc{Qwen2.5-7B-Instruct} on ECQA (Appendix~\ref{apx:cross-family}). The trends hold: DPO improves worst-case Spearman (10.83$\rightarrow$10.87) while preserving higher-ranked explanations (36.98$\rightarrow$37.02), and cosine similarity improves across all ranks (e.g., worst: 5.55$\rightarrow$5.57, best: 6.48$\rightarrow$6.61). As with LLaMA, the strongest gains target the weakest explanations, suggesting attribution-guided preference optimization captures an architecturally general pattern.

\paragraph{SFT Baseline Comparison.}
To isolate the role of the contrastive training signal in DPO, we compare against a supervised fine-tuning (SFT) baseline trained on the same highest-ranked explanations that serve as the chosen examples in DPO, without exposure to rejected explanations.
Table~\ref{tab:sft_vs_dpo} reports results on ECQA for LLaMA3.1-8B with LIME attributions.
SFT substantially degrades Spearman correlation in both the T--T and B--T settings: mean CC-Sp drops by 41.4\% and 37.3\%, respectively, relative to the B--B baseline.
Cosine similarity also decreases in the B--T setting ($-$6.6\%).
In contrast, DPO improves Spearman correlation by 13.0\% (T--T) and 11.0\% (B--T), with cosine gains of up to 20.8\%.
These results demonstrate that simply maximizing the likelihood of high-consistency explanations is insufficient for improving self-consistency; the contrastive signal provided by DPO, comparing preferred and rejected explanations, is essential for guiding the model toward more aligned reasoning.

\paragraph{Cross-Metric Transfer.}
Although training is guided only by the Spearman correlation, we observe improvements in cosine similarity across both LIME and LIG. This cross-metric transfer suggests that optimizing for rank alignment not only stabilizes feature prioritization but also enhances directional attribution similarity. Notably, for LLaMA3.2-3B on ECQA, cosine similarity improvements (up to \textit{92.8\%}) exceed Spearman gains, indicating positive spillover effects. The consistency of these results across attribution methods further supports the robustness of our approach.

\subsection{Cross-Domain Generalization}
Table~\ref{tab:cross_domain_eval} reports cross-domain performance across evaluation modes.
In the \textbf{T--T mode}, models fine-tuned on ECQA or ARC-Easy show consistent improvements in Spearman correlation.
For LLaMA3.1-8B, gains range from 10.9--13.9\%, with the strongest transfer from ARC-Easy to the more challenging ARC-Challenge (+13.9\%).
The smaller LLaMA3.2-3B also benefits, with gains between 9.5--14.3\%.
Cosine similarity shows a more moderate but still positive transfer: for LLaMA3.1-8B, improvements range from 4.1--6.4\%, while for LLaMA3.2-3B, gains span 5.8--18.3\%, including a notable improvement when transferring from ARC-Easy to CODAH.
Complementary results in the \textbf{B--T mode} show weaker and sometimes mixed trends, indicating that the strongest gains arise when both the decision and explanation are generated by the tuned model in the T--T setting. In general, these findings suggest that our DPO-based approach encourages explanation strategies that generalize beyond the training domain.

\subsection{Plausibility and Truthfulness Evaluation}

Because self-consistency is defined in terms of internal attributional alignment between a model's decision and its explanation, it cannot be directly assessed by human annotators.
Accordingly, our primary evaluation of self-consistency relies on attribution-based measures.
To verify that optimizing for self-consistency does not degrade user-facing explanation quality, we conduct a small-scale human plausibility study as a sanity check (Appendix~\ref{apx:plaus_study}). This study is intentionally scoped to assess surface-level quality rather than attributional faithfulness, which is inherently inaccessible to human annotators.

Three annotators evaluated 80 explanations sampled across all models and datasets, judging (i) whether each explanation was expressed in natural language and (ii) its surface-level plausibility on a 3-point scale.
Nearly all explanations were judged to be natural language (98.8\% for base models and 100\% for DPO-tuned models), with average plausibility scores of $2.92/3.0$ and $2.88/3.0$, respectively.
These results indicate that DPO fine-tuning preserves the naturalness and plausibility of explanations, and does not introduce artifacts that would reduce human interpretability.

We further evaluate factual reliability by testing both vanilla and DPO-tuned models on \textsc{TruthfulQA}~\citep{lin2021truthfulqa}, which probes a model's tendency to produce false or misleading answers under adversarial questioning. DPO-tuned models achieve comparable or slightly higher accuracy across both variants (Appendix~\ref{apx:truthfulqa}), showing that improving attributional self-consistency does not compromise factual accuracy.

\subsection{Self-Consistency is Multidimensional}
\label{sec:cross_method}

A key finding of this work is that attribution-based self-consistency is not a single unified property but a multidimensional one.
Notably, while our method generalizes robustly across domains (Section~\ref{sec:in_domain_eval}), it does not transfer across attribution paradigms.
Models fine-tuned using LIME-based alignment are evaluated with LIG-based metrics and vice versa,
with additional evaluation using the sampling-based \textsc{KSHAP}~\citep{lundberg2017unified} method.
Results are reported in Appendix~\ref{apx:cross-method}.
Across all settings, improvements largely remain within the training method,
confirming that each attribution approach captures a fundamentally distinct notion of feature importance.

This outcome is consistent with the theoretical differences between these paradigms: LIME fits local linear surrogates via input perturbation, while Integrated Gradients accumulates sensitivity along a baseline path, and KSHAP estimates Shapley values through sampling. These methods encode different assumptions about how to decompose a prediction into token-level contributions, and prior comparative analyses confirm they often disagree in practice~\citep{Krishna2022TheDP,Neely2021OrderIT,Neely2022ASO}.
To quantify this, we compare how LIME and LIG rank the same set of explanations across 10{,}880 ECQA examples:
their rankings exhibit only moderate agreement, rarely sharing the same top explanation (Top-1~$\approx0.2$) but often overlapping among the top three (Top-3~$\approx0.6$).

Importantly, this does not indicate that our optimization exploits method-specific artifacts: although training uses only Spearman, the unoptimized cosine metric also improves strongly (up to 92.8\% on ECQA, Section~\ref{sec:in_domain_eval}). Gaming method-specific noise would degrade unoptimized metrics, not improve them.

These findings carry broader implications for the attribution literature: practitioners should select the attribution method whose assumptions best match their evaluation goals, and our framework provides a general-purpose pipeline for optimizing self-consistency under any chosen definition of input relevance.
\section{Conclusion}
We introduce the \emph{PSCB}, enabling the first large-scale study of how closely LLM explanations align with the evidence behind their decisions. Our analysis shows that \emph{self-consistency} represents a distinct and interpretable aspect of model behavior, largely independent of answer correctness. Rank-based Spearman correlation offers a sharper and more stable signal than cosine for assessing this alignment. Leveraging it via preference supervision yields consistent improvements in self-consistency across models, datasets, and attribution methods. The strongest gains appear when the same tuned model generates both decisions and explanations, indicating that shared representations enhance consistency. These improvements come without compromising task accuracy, natural language form, plausibility, or truthfulness. Overall, our results establish the first scalable framework for quantifying and improving attributional self-consistency, advancing language models whose explanations are grounded in the same input evidence that drives their decisions.

\paragraph{Future Directions.}
Natural extensions include adapting the framework to open-ended generation by aggregating attributions over answer spans, jointly optimizing across multiple attribution paradigms given that they capture complementary notions of importance (Section~\ref{sec:cross_method}), and an online variant that recomputes attributions during training to close the residual gap between the tuned model and the offline attribution signal it was optimized against.
A complementary direction is testing whether higher self-consistency yields measurable user gains, e.g., better-calibrated trust.
\section*{Limitations}
This work focuses on assessing and improving the
self-consistency of natural language explanations
given by LLMs. We report the following limitations:

\paragraph{Attribution Methods and Scalability.}
Attribution-based self-consistency analysis remains computationally demanding, which limits the range of methods that can be feasibly applied at scale. While theoretically principled approaches such as CC-SHAP~\citep{parcalabescu2023measuring} provide strong guarantees, they require more than 4 minutes per example, rendering large-scale benchmarking and preference-based training impractical. To balance rigor and scalability, our main experiments rely on the perturbation-based LIME method~\citep{ribeiro2016should}, complemented by the gradient-based Layer Integrated Gradients (LIG) approach~\citep{sundararajan2017axiomatic}. These choices enable consistent evaluation across tens of thousands of examples while still capturing complementary perspectives on feature importance.

As shown in Section~\ref{sec:cross_method}, improvements under one attribution paradigm do not generalize to others, a limitation inherent to all attribution-based evaluations. This highlights both the difficulty of defining a universal measure of explanation faithfulness and the value of our framework for optimizing under any chosen attribution method.

\paragraph{Human Evaluation.}
Because attributional self-consistency depends on internal model signals inaccessible to human annotators, our primary evaluation is attribution-based. The plausibility study (Section~\ref{sec:empirical_evaluation}) serves only as a sanity check on surface-level quality.
Whether increased self-consistency translates into more \emph{helpful} explanations from a user perspective remains an open question, and we view systematic human studies of downstream utility as an important direction for future work.

\paragraph{Model and Task Scope.}
Our study focuses on small to medium scale instruction-tuned LLMs evaluated in a multiple-choice question answering (MCQA) setting.
We adopt MCQA as a controlled testbed for studying attributional self-consistency, as it provides a well-defined decision signal and enables reliable comparison between decision and explanation attributions across models.
This design choice prioritizes attributional reliability and experimental control over task generality.

The proposed optimization framework itself is not tied to MCQA and can in principle be applied to open-ended decision settings, including free-form reasoning or summarization.
However, extending PSCB-style attribution benchmarks to such tasks substantially increases attribution and evaluation complexity.
While we include a lightweight cross-model-family validation, a systematic evaluation across diverse architectures and open-ended tasks remains an important direction for future work.

\paragraph{Offline vs. Online.}
To improve the LLM's self-consistency, we update it using DPO in an offline manner, creating a new model dedicated to providing consistent explanations. However, this approach introduces a trade-off: while the updated model generates more self-consistent explanations, it achieves only approximate self-consistency since it differs from the original model that generated the reference answers. Online learning could theoretically maintain perfect self-consistency by updating the model while continuously re-calculating attribution vectors for its own evolving outputs. However, this approach presents significant technical challenges, including the computational overhead of recalculating the attributions at each training step and the potential instability in the learning dynamics, making offline training the more practical option despite its inherent approximation.

\section*{Acknowledgments}
Funded by the European Union (ERC, Convey, 101078158). Views and opinions expressed are however those of the author(s) only and do not necessarily reflect those of the European Union or the European Research Council Executive Agency. Neither the European Union nor the granting authority can be held responsible for them.

\bibliography{bibs}

\appendix
\label{sec:appendix}
\section{Prompt Templates}
\label{apx:prompt-templates}

\paragraph{Decision Prompt.}
To elicit the model’s answer:

\begin{quote}
\ttfamily
Question: \{question text\} \\
Choose the most plausible answer, respond only with the answer and the description:

A. \{option A\} \\
B. \{option B\} \\
C. \{option C\} \\
D. \{option D\} \\

Answer:
\end{quote}

The model is expected to respond with a single letter corresponding to its chosen option and the answer content (e.g., “A. {option A}”).

\paragraph{Explanation Prompt.}
To generate a justification for the selected answer, we prompt the model as follows:

\begin{quote}
\ttfamily
Question: \{question text\} \\
Choose the most plausible answer, respond only with the answer and the description:

A. \{option A\} \\
B. \{option B\} \\
C. \{option C\} \\
D. \{option D\} \\

Selected Answer: \{model's answer\}

Why did you make that choice? Explain briefly.

Explanation: \{model's explanation\}
\end{quote}

The explanation should rely only on the information provided in the question and answer choices.

\section{Implementation Details}
\label{apx:implementation_details}

\paragraph{Environment.} All experiments were conducted using NVIDIA A100 80GB GPUs. We used PyTorch and Hugging Face Transformers. Captum was used for attribution computations. All models were accessed through the Hugging Face Hub and run in float16 mode.

\paragraph{Dataset Processing.} Each multiple-choice QA dataset (ECQA, ARC-Easy, ARC-Challenge, CODAH) was converted into a unified format consisting of the question, 4–5 answer choices, the model’s predicted answer, and $k=5$ sampled post-hoc explanations. All datasets were split into training (70\%), validation (20\%), and test (10\%) sets using a fixed random seed (42). We precomputed attribution vectors for the model's decision and each explanation, resulting in six LIME runs per scenario.

\paragraph{Prompting.} Decisions were elicited using the following template: \texttt{“Choose the most plausible answer:”}, followed by the answer options. Explanations were generated using the model's answer with the prompt: \texttt{“Why did you make that choice? Explain briefly.”} (see Appendix~\ref{apx:prompt-templates} for full prompt templates).

\paragraph{Text Generation.} We used nucleus sampling with top-$p = 0.9$, temperature = 0.7, and a maximum generation length of 400 tokens. Generation was done in a zero-shot setting. Padding tokens were manually set to a new padding token for compatibility. For each input, we generated 5 explanations using fixed random seeds ranging from 42 to 46 to ensure diversity and reproducibility.

\paragraph{Attribution.} Feature attributions were computed using LIME with 500 perturbation samples per example, using Captum’s implementation. The reference input consisted of the pad token repeated to the input length. A manually defined list of formatting and punctuation tokens was excluded from attribution (see Appendix~\ref{apx:skip_tokens_list}).

\paragraph{Consistency Metrics.} We computed cosine similarity and Spearman rank correlation between the attribution vectors of the model’s decision and each explanation. These scores were used to rank explanations and construct attribution-based preference pairs.

\paragraph{SFT Baseline.} To evaluate whether a standard supervised fine-tuning objective can improve self-consistency, we train an SFT baseline on the same highest-ranked explanations used as the chosen examples in DPO. The SFT model is fine-tuned to maximize the likelihood of these preferred explanations using the same LoRA configuration, training epochs, batch size, and optimizer as the DPO models. The learning rate is $6.95 \times 10^{-6}$. This isolates the effect of the contrastive DPO signal by removing the rejected explanation from the training procedure.

\paragraph{DPO Fine-Tuning.} We used Direct Preference Optimization (DPO) to fine-tune each model using preference pairs derived from attribution alignment scores. We sampled 5 explanations per instance, ranked them using Spearman correlation, and used the highest- and lowest-ranked as the preferred and rejected explanations, respectively. Fine-tuning was done with LoRA for parameter-efficient updates. We apply LoRA with a rank of 32 and scaling factor \texttt{lora\_alpha} = 32, targeting all major projection layers (\texttt{q\_proj}, \texttt{k\_proj}, \texttt{v\_proj}, \texttt{o\_proj}, \texttt{gate\_proj}, \texttt{up\_proj}, \texttt{down\_proj}) and disabling dropout and bias adaptation. Gradient checkpointing is enabled via the \texttt{unsloth} backend for memory efficiency. All models were fine-tuned independently per dataset. We summarize the hyperparameters used for fine-tuning all models in Table~\ref{tab:dpo_hyperparams}.

\begin{table*}[htbp]
    \centering
    \small
    \begin{tabular}{lcccc}
        \toprule
        \textbf{Hyperparameter} & ECQA (L3.1) & ARC-Easy (L3.1) & ECQA (L3.2) & ARC-Easy (L3.2) \\
        \midrule
        Epochs                & 10                 & 10                 & 10                 & 10                 \\
        Batch Size            & 16                & 16                & 16                & 16                \\
        Gradient Accumulation            & 8                & 8                & 8                & 8                \\
        Learning Rate         & $4.21 \times 10^{-6}$ & $4.65 \times 10^{-6}$ & $ 9.55 \times 10^{-6}$ & $6.32 \times 10^{-6}$ \\
        DPO Beta              & 5.13               & 5.64               & 8.44               & 8.84               \\
        Score Scale Factor              & 10               & 10               & 10               & 10               \\
        Optimizer             & AdamW             & AdamW             & AdamW             & AdamW             \\
        \bottomrule
    \end{tabular}
    \caption{Hyperparameters used for DPO fine-tuning across model–dataset pairs.}
    \label{tab:dpo_hyperparams}
\end{table*}

\section{Skip Tokens for Attribution}
\label{apx:skip_tokens_list}

To ensure attribution focuses on semantically meaningful input content, we exclude formatting and structure-related tokens from all attribution computations. For LLaMA models, we identify a set of \textit{skip tokens} that should not be considered when measuring input importance. These tokens include both model-specific structural markers and general-purpose formatting symbols.

\paragraph{Structure Tokens.} Based on the tokenizer vocabulary and architecture of \textbf{LLaMA3.1} and \textbf{LLaMA3.2}, we exclude the following structure tokens when present:

\begin{itemize}
    \item \texttt{<|start\_header\_id|>}
    \item \texttt{<|end\_header\_id|>}
    \item \texttt{<|eot\_id|>}
    \item \texttt{<|begin\_of\_text|>}
    \item \texttt{Ċ} (newline marker or formatting artifact)
    \item \texttt{Ġ->} (common artifact from tokenizer for arrows or prompt delimiters)
\end{itemize}

These tokens typically serve as formatting scaffolding or internal delimiters in system and chat prompts, and do not reflect actual semantic content from the question or explanation.

\paragraph{Usage.} We apply this skip list to both the decision and explanation inputs before computing attribution scores. Tokens that match the above set (by string or token ID) are held fixed during perturbation and excluded from similarity calculations between attribution vectors.

\paragraph{Note.} We do not exclude standard stop words or punctuation in our main experiments, as their contribution may still reflect the model's learned reasoning behavior. However, our framework allows toggling this behavior for ablation studies.

\section{DPO vs.\ Vanilla Example}
\label{apx:dpo_example}

\begin{figure*}[htbp]
    \centering
    \includegraphics[width=0.9\linewidth]{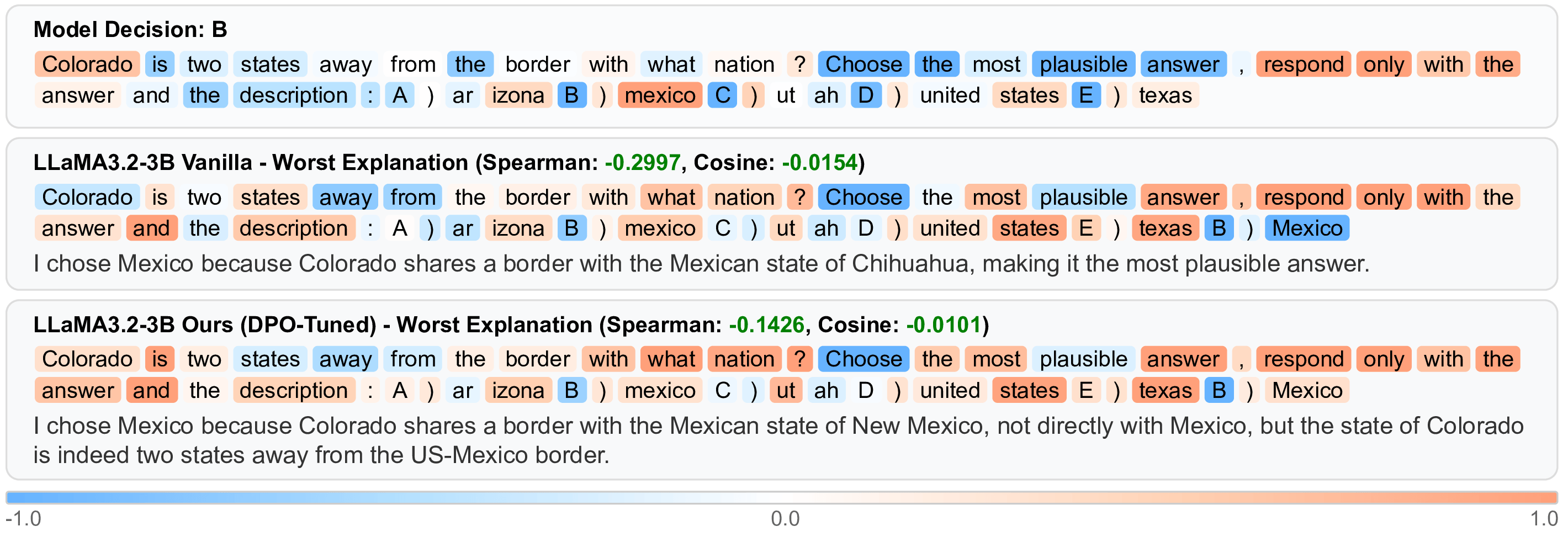}
    \caption{An example from ECQA shows attribution alignment for the LLaMA3.2-3B model's decision and its worst explanation across two variants. The model selects ``B'' (Mexico), with high-attribution tokens like ``Colorado,'' ``two,'' ``states,'' and ``Mexico'' supporting its geographic reasoning. The DPO-tuned model generates a worst-case explanation that better reflects this rationale, noting Colorado's distance from Mexico, mirroring the decision attribution. In contrast, the vanilla model incorrectly references a border with ``Chihuahua,'' unsupported by the decision attribution.}
    \label{fig:example3}
\end{figure*}

\section{Qualitative Examples Heatmaps}
\begin{figure*}[htbp]
    \centering

    \begin{subfigure}{\linewidth}
        \centering
        \includegraphics[width=\linewidth]{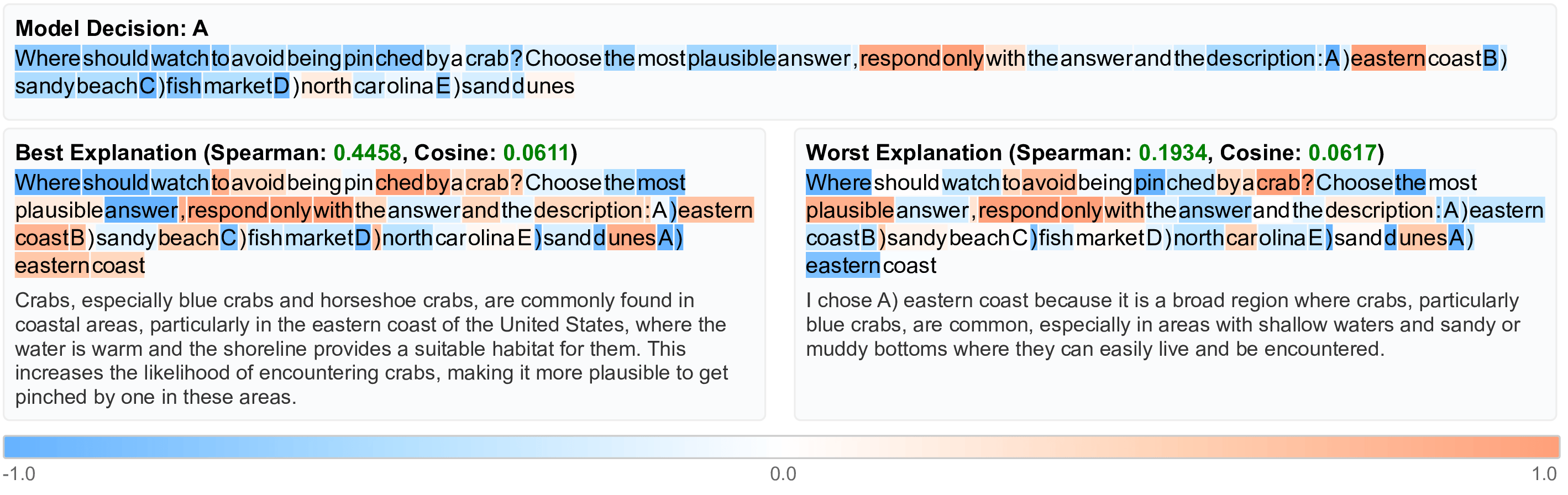}
        \caption{Vanilla LLaMA3.1, ECQA, LIME}
        \label{fig:example6}
    \end{subfigure}

    \vspace{0.5em} 

    \begin{subfigure}{\linewidth}
        \centering
        \includegraphics[width=\linewidth]{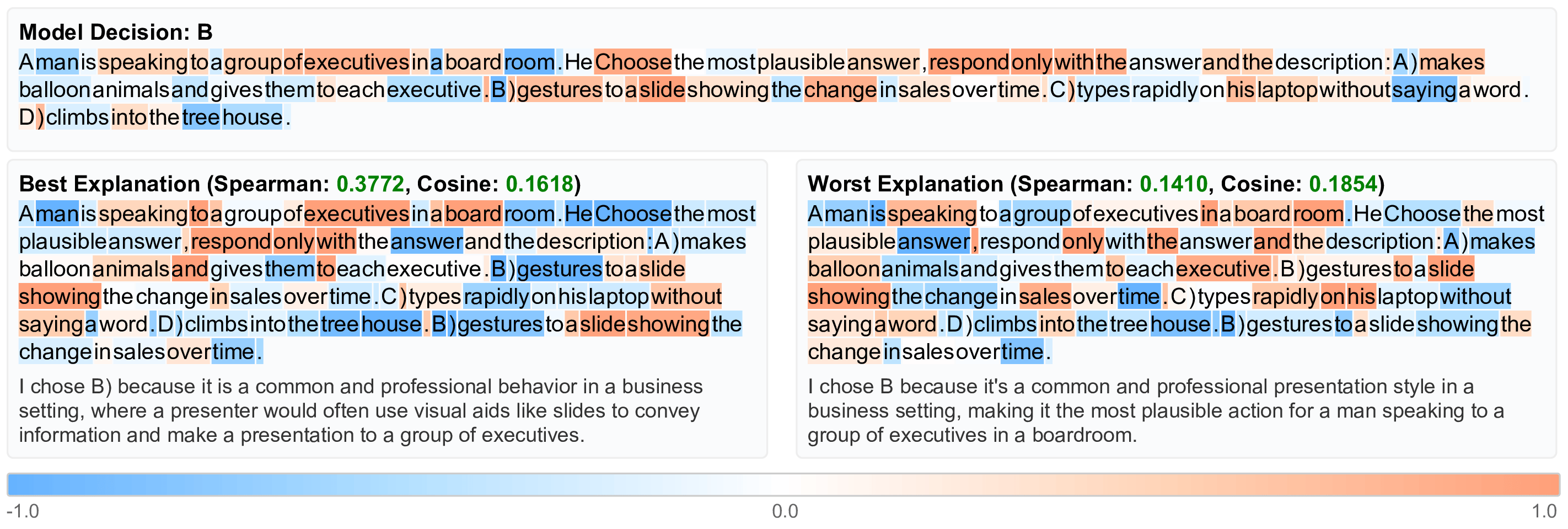}
        \caption{Vanilla LLaMA3.1, ECQA, LIME}
        \label{fig:example7}
    \end{subfigure}

    \vspace{0.5em}

    \begin{subfigure}{\linewidth}
        \centering
        \includegraphics[width=\linewidth]{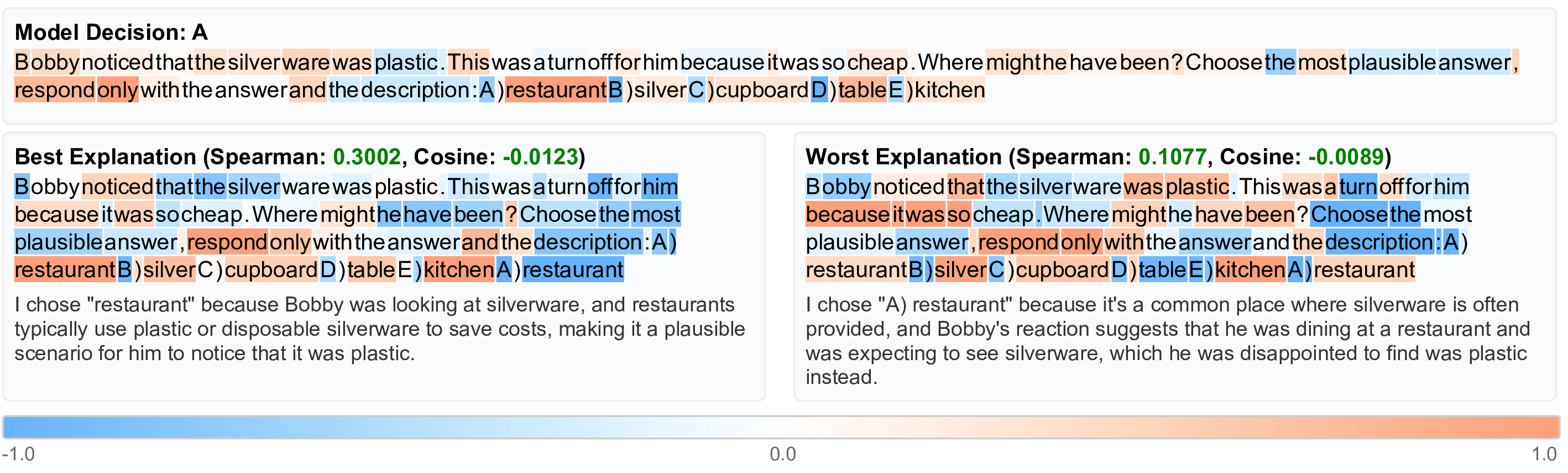}
        \caption{Vanilla LLaMA3.2, ECQA, LIME}
        \label{fig:example8}
    \end{subfigure}

    \caption{Qualitative heatmap visualizations showing token-level attribution scores for model decisions and explanations using the LIME method.}
    \label{fig:qualitative_heatmaps_vanilla}
\end{figure*}

Figure~\ref{fig:qualitative_heatmaps_vanilla} presents token-level attribution heatmaps for model decisions and explanations, using the LIME method. Each subfigure illustrates a different scenario from the ECQA dataset, comparing the best and worst explanations produced by vanilla models based on Spearman and Cosine alignment scores. Across examples, we observe that high-quality explanations (left) tend to emphasize tokens that align more closely with the model’s decision rationale, for instance, highlighting location-specific cues like “eastern coast” or situational context like “board room.” In contrast, low-ranking explanations (right) often shift focus to semantically irrelevant or misleading tokens, despite sounding plausible. These visualizations underscore the discriminative power of attribution-based metrics and highlight the variability in explanation quality for the same input.

\section{TruthfulQA Evaluation}
\label{apx:truthfulqa}

We further evaluate whether optimizing for self-consistency affects factual reliability using the TruthfulQA benchmark~\cite{lin2021truthfulqa}. 
Table~\ref{apx:tab_truthfulqa} reports multiple-choice accuracy for the vanilla and DPO-tuned models under both LIME- and LIG-based consistency training. 
Across all settings, accuracy differences remain marginal ($<1\%$), confirming that improving attributional self-consistency does not compromise factual correctness. 
In some cases (e.g., ARC-Easy–trained L3.1–8B), the DPO-tuned models slightly outperform their vanilla counterparts, suggesting that the enhanced alignment between decisions and explanations may even reinforce truthful reasoning.
These results support that our optimization procedure preserves factual accuracy while improving internal consistency.

\begin{table}[ht]
\centering
\small
\begin{tabularx}{\columnwidth}{l l l Y Y}
\toprule
& \textbf{Model} & \textbf{Training Data} & \textbf{Acc. (\%)} & \textbf{vs Vanilla} \\
\midrule
\multirow{6}{*}{\rotatebox[origin=c]{90}{LIME}} 
  & \multirow{3}{*}{L3.1-8B} 
    & Vanilla   & 53.86  & --   \\
  &           & ECQA     & 53.73  & -0.13 \\
  &           & ARC-Easy & 54.71  & +0.85 \\
\cmidrule(lr){2-5}
  & \multirow{3}{*}{L3.2-3B} 
    & Vanilla   & 52.75  & --   \\
  &           & ECQA     & 52.63  & -0.12 \\
  &           & ARC-Easy & 53.0  & +0.25 \\
\midrule
\multirow{6}{*}{\rotatebox[origin=c]{90}{LIG}} 
  & \multirow{3}{*}{L3.1-8B} 
    & Vanilla   & 53.86  & --   \\
  &           & ECQA     & 54.10  & +0.24 \\
  &           & ARC-Easy & 53.98  & +0.10 \\
\cmidrule(lr){2-5}
  & \multirow{3}{*}{L3.2-3B} 
    & Vanilla   & 52.75  & --   \\
  &           & ECQA     & 53.49  & +0.74 \\
  &           & ARC-Easy & 53.12  & +0.37 \\
\bottomrule
\end{tabularx}
\caption{TruthfulQA MCQA accuracy as a factual reliability check. Higher is better.}
\label{apx:tab_truthfulqa}
\end{table}

\section{Cross-Method Generalization}
\label{apx:cross-method}

We further assess whether improvements in self-consistency transfer across attribution methods. 
Specifically, we train models using one attribution framework (e.g., LIG) and evaluate them using another (e.g., LIME), as shown in Table~\ref{apx:tab_cross_method}. 
This cross-method setup tests whether the model internalizes a general notion of alignment. 
Results indicate that DPO-tuned models exhibit limited transfer: improvements in one attribution method do not reliably extend to another. 
For example, models trained with LIG show negligible or inconsistent gains when evaluated with LIME metrics, and vice versa. 
This outcome aligns with recent evidence that different attribution paradigms capture complementary aspects of model reasoning rather than identical importance structures. 


\begin{table}[ht]
\centering
\small
\begin{tabularx}{\columnwidth}{l l l | c | Y | Y}
\toprule
& & \textbf{Training} &
\textbf{Acc.} &
\textbf{CC-Cos~\scriptsize($\uparrow\downarrow$)} & \textbf{CC-Sp~\scriptsize($\uparrow\downarrow$)} \\
\midrule
\multirow{6}{*}{\rotatebox[origin=c]{90}{\textbf{ECQA}}} & \multirow{3}{*}{\rotatebox[origin=c]{90}{L3.1-8B}}
& None & 68.40 & \textbf{09.57} $\pm$ 0.31 & 17.68 $\pm$ 0.47 \\
&& ECQA & 71.20 & 09.24 $\pm$ 0.32 & 17.73 $\pm$ 0.48 \\
&& ARC-E & 70.00 & 09.28 $\pm$ 0.33 & \textbf{18.08} $\pm$ 0.43 \\
\cmidrule{2-6}
& \multirow{3}{*}{\rotatebox[origin=c]{90}{L3.2-3B}}
& None & 68.00 & 0.33 $\pm$ 0.30 & \textbf{19.36} $\pm$ 0.57 \\
&& ECQA & 67.20 & 0.46 $\pm$ 0.31 & 19.23 $\pm$ 0.56 \\
&& ARC-E & 66.80 & \textbf{0.57} $\pm$ 0.30 & 19.27 $\pm$ 0.57 \\
\midrule

\multirow{6}{*}{\rotatebox[origin=c]{90}{\textbf{ARC-Easy}}} & \multirow{3}{*}{\rotatebox[origin=c]{90}{L3.1-8B}}
& None & 83.20 & 15.56 $\pm$ 0.34 & \textbf{13.65} $\pm$ 0.49 \\
&& ECQA & 82.80 & \textbf{15.73} $\pm$ 0.36 & 13.17 $\pm$ 0.52 \\
&& ARC-E & 82.80 & 15.51 $\pm$ 0.35 & 13.51 $\pm$ 0.52 \\
\cmidrule{2-6}
& \multirow{3}{*}{\rotatebox[origin=c]{90}{L3.2-3B}}
& None & 76.80 & 06.48 $\pm$ 0.44 & 17.30 $\pm$ 0.54 \\
&& ECQA & 76.80 & \textbf{06.49} $\pm$ 0.44 & 17.33 $\pm$ 0.55 \\
&& ARC-E & 76.80 & 06.42 $\pm$ 0.44 & \textbf{17.45} $\pm$ 0.56 \\
\bottomrule
\end{tabularx}
\caption{\textbf{cross-method generalization: LIG model using LIME evaluations}  
Values aggregate over 5 explanations per item (Mean). 
$\Delta$ is DPO - Vanilla.}
\label{apx:tab_cross_method}
\end{table}

\section{Plausibility Study}
\label{apx:plaus_study}

To confirm that optimizing for self-consistency did not compromise the readability of explanations, we conducted a small-scale plausibility study. 

\paragraph{Participants.} Three graduate students participated in the study. 

\paragraph{Materials.} We sampled 80 explanations in total: 5 examples from each condition defined by attribution method (LIME, LIG), dataset (ECQA, ARC-Easy), model (LLaMA3.1, LLaMA3.2), and variant (Base, DPO-Tuned). Each explanation was presented as full text together with the model’s selected decision.

\paragraph{Procedure.} The study was administered via Qualtrics, with questions randomized. For each explanation, annotators evaluated two criteria:
(i) whether the explanation was expressed in fluent natural language (binary yes/no), and
(ii) how plausible the explanation appeared on a 3-point Likert scale (1 = implausible, 3 = highly plausible).
Annotators were instructed not to consider alternative answers, external knowledge, or internal model reasoning.
No comparison between explanations, ranking tasks, or faithfulness judgments were requested.

\paragraph{Results.} Nearly all explanations were judged to be natural language (98.8\% for base models and 100\% for DPO-tuned models). The mean plausibility scores were $2.92/3.0$ for base models and $2.88/3.0$ for DPO-tuned models. Annotators showed high percent agreement across both criteria, with ratings broadly consistent across participants. These results confirm that explanations remain coherent and generally human-plausible, even after DPO fine-tuning. We emphasize that this study is intentionally scoped as a sanity check on surface-level explanation quality, rather than a validation of attributional faithfulness, which depends on internal model signals inaccessible to human annotators.

\paragraph{Example Shown}
\noindent\texttt{Scenario: What is the capital of France?  
A) Rome \quad B) Paris \quad C) Madrid \quad D) Berlin \\  
Model's Decision: B) Paris}

\begin{figure}[t]
    \centering
    \includegraphics[width=\columnwidth]{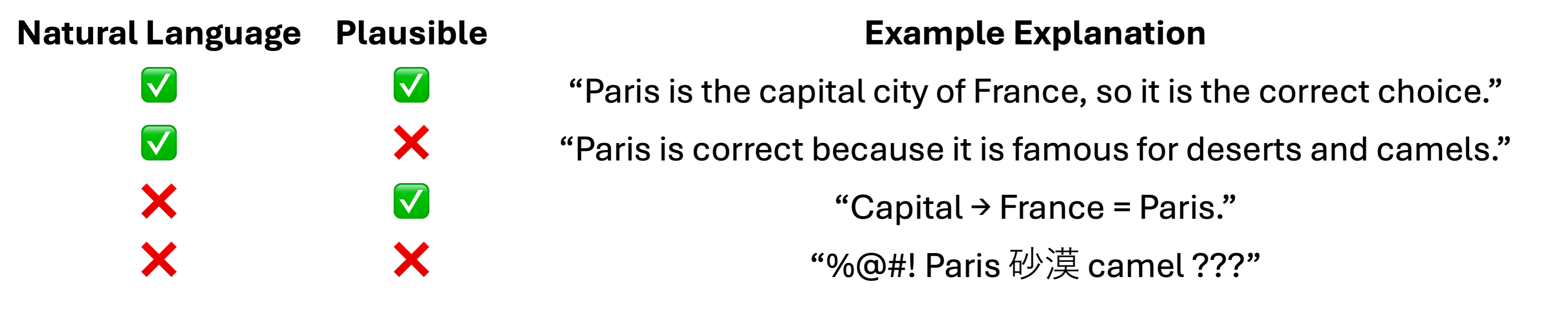}
    \caption{Illustrated example shown to participants during annotation.}
    \label{fig:intro}
\end{figure}

\section{Cross-Model-Family Validation}
\label{apx:cross-family}

To examine whether the observed self-consistency improvements are tied to the LLaMA architecture, we perform a lightweight cross-model-family validation on the \textsc{ECQA} dataset using \textsc{Qwen2.5-7B-Instruct}.
ECQA offers a controlled multiple-choice setting with a well-defined decision signal, allowing us to apply the same attribution, alignment, and evaluation pipeline used in the main experiments without modification.

\paragraph{Setup.}
We evaluate \textsc{Qwen2.5-7B-Instruct} under the same Decider–Explainer configurations employed throughout the paper, comparing a vanilla model against a DPO-tuned variant trained using attribution-based preferences.
Due to the substantial computational cost of constructing a full PSCB benchmark for an additional model family, this validation is restricted to ECQA.
Unlike prior small-scale attribution studies, we nonetheless use the full ECQA test set rather than a reduced subset.
Self-consistency is measured primarily using Spearman rank correlation, with cosine similarity reported as a secondary metric.
All attribution methods, hyperparameters, and evaluation procedures are identical to those used for the LLaMA models, and all scores are reported on the same scale as in the main results.

\paragraph{Results.}
The results follow the same qualitative trends observed for the LLaMA family.
Under Spearman correlation, DPO tuning improves self-consistency at the lower end of the explanation quality distribution, increasing the worst-case score from 10.829 to 10.871, while leaving the best-case score largely unchanged (36.980 to 37.021).
The median Spearman score decreases slightly from 24.163 to 23.235, reflecting the same redistribution pattern seen in earlier experiments.
Cosine similarity shows consistent improvements across all ranks, with increases in the worst (5.547 to 5.571), median (5.964 to 6.080), and best (6.478 to 6.607) explanations.
As in the LLaMA experiments, the most reliable gains occur for the weakest explanations, while changes to higher-quality explanations remain modest.

\paragraph{Discussion.}
Although the absolute magnitude of the improvements is smaller than for the LLaMA models, the qualitative behavior closely mirrors the central findings of the paper.
Attribution-guided preference optimization primarily strengthens low-quality explanations while largely preserving stronger ones, consistent with the intended objective of the method.
This cross-model-family validation therefore provides evidence that the proposed framework captures an optimization pattern that is not specific to a single architecture, within the same operational definition of attributional self-consistency.
We stress that this experiment is intentionally limited in scope and is intended as a sanity check rather than a replacement for a full multi-family PSCB construction.

\end{document}